\documentclass{article}

\usepackage[nonatbib,final]{neurips_2023}

\usepackage{amsmath,amsfonts,bm}

\def\eqref#1{equation~\ref{#1}}

\def\1{\bm{1}}

\DeclareMathAlphabet{\mathsfit}{\encodingdefault}{\sfdefault}{m}{sl}
\SetMathAlphabet{\mathsfit}{bold}{\encodingdefault}{\sfdefault}{bx}{n}

\usepackage[utf8]{inputenc} 
\usepackage[T1]{fontenc}    
\usepackage{hyperref}       
\usepackage{url}            
\usepackage{booktabs}       
\usepackage{amsfonts}       
\usepackage{nicefrac}       
\usepackage{microtype}      
\usepackage{xcolor}         
\usepackage{multirow}
\usepackage[normalem]{ulem}
\usepackage{graphicx}
\usepackage{authblk}
\usepackage{makecell}
\usepackage{wrapfig}
\usepackage{tabularray}

\title{Making LLaMA SEE and Draw with SEED Tokenizer}
\vspace{-10pt}
\begin{document}

\author{
\textbf{Yuying Ge$^{1\ast}$ \qquad Sijie Zhao$^{1\ast}$ \qquad Ziyun Zeng$^{2}$ \qquad Yixiao Ge$^{1,2\dagger}$\\} \textbf{\qquad Chen Li$^{2}$ \qquad Xintao Wang$^{1,2}$ \qquad Ying Shan$^{1,2}$}

$^{1}$Tencent AI Lab \qquad $^{2}$ARC Lab, Tencent PCG

\url{https://github.com/AILab-CVC/SEED}
}

\maketitle

\begin{figure}[h!]
\vspace{-30pt}
	\centering
	\includegraphics[width=0.92\linewidth]{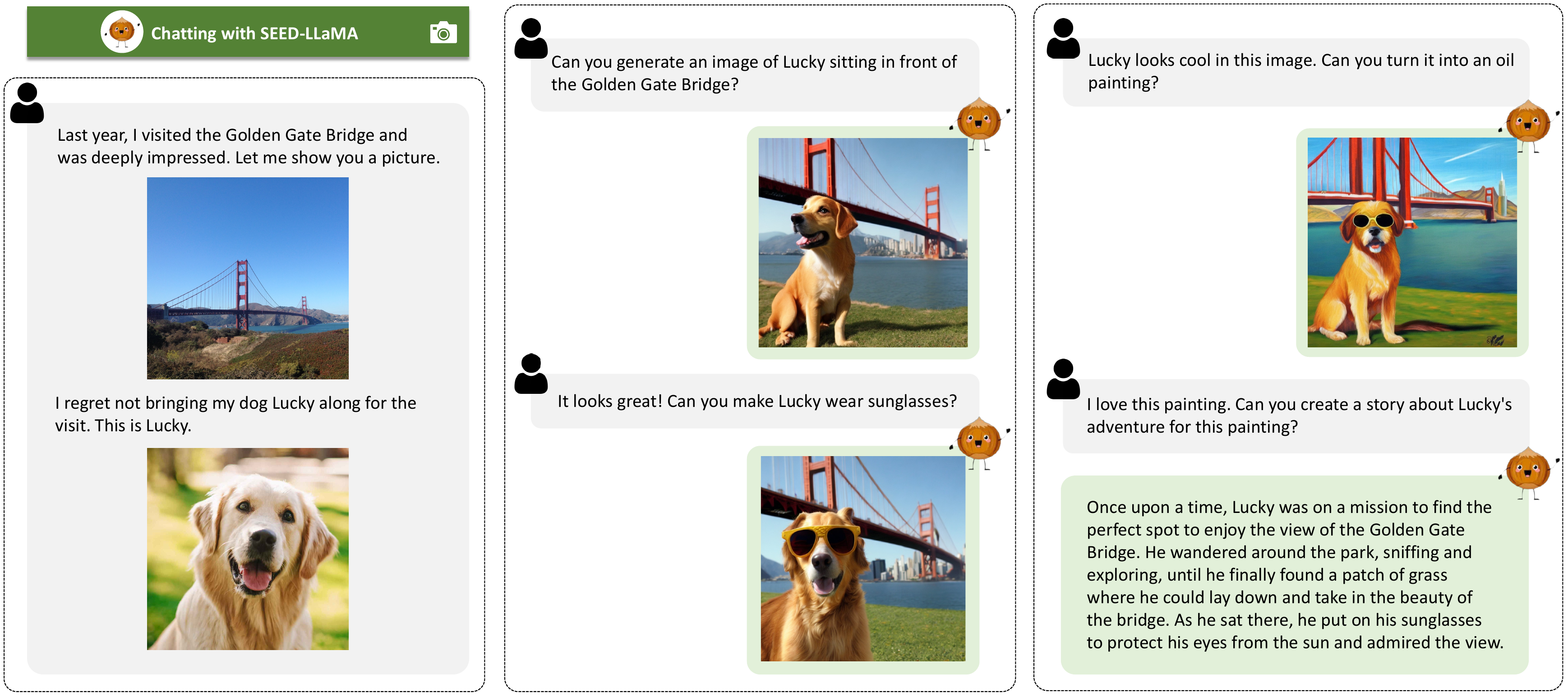}
 \vspace{-5pt}
\caption{The introduced SEED-LLaMA, a multimodal AI assistant, demonstrates \textbf{emergent ability} in the multi-turn in-context image and text generation given multimodal instructions.}
\vspace{-5pt}
	\label{fig:teaser_example}
\end{figure}

\begin{abstract}
\vspace{-5pt}
The great success of Large Language Models (LLMs) has expanded the potential of multimodality, contributing to the gradual evolution of General Artificial Intelligence (AGI). A true AGI agent should not only possess the capability to perform predefined multi-tasks but also exhibit emergent abilities in an open-world context. However, despite the considerable advancements made by recent multimodal LLMs, they still fall short in effectively unifying comprehension and generation tasks, let alone open-world emergent abilities. We contend that the key to overcoming the present impasse lies in enabling text and images to be represented and processed interchangeably within a unified autoregressive Transformer. To this end, we introduce \textbf{SEED}, an elaborate image tokenizer that empowers LLMs with the ability to \textbf{SEE} and \textbf{D}raw at the same time. We identify two crucial design principles:
(1) Image tokens should be independent of 2D physical patch positions and instead be produced with a \textit{1D causal dependency}, exhibiting intrinsic interdependence that aligns with the left-to-right autoregressive prediction mechanism in LLMs. (2) Image tokens should capture \textit{high-level semantics} consistent with the degree of semantic abstraction in words, and be optimized for both discriminativeness and reconstruction during the tokenizer training phase. With SEED tokens, LLM is able to perform scalable multimodal autoregression under its original training recipe, {i.e.}, next-word prediction. SEED-LLaMA\footnote{This work is a follow-up of SEED~\cite{ge2023planting}, where we update the visual tokenizer and present SEED-LLaMA.} is therefore produced by large-scale pretraining and instruction tuning on the interleaved textual and visual data, demonstrating impressive performance on a broad range of multimodal comprehension and generation tasks. More importantly, SEED-LLaMA has exhibited compositional emergent abilities such as multi-turn in-context multimodal generation, acting like your AI assistant.

\end{abstract}

\section{Introduction}
In recent years, Large Language Models~\cite{touvron2023llama, brown2020language, chowdhery2022palm} (LLMs) pre-trained on massive text corpus with straightforward training objectives such as next-word prediction have exhibited remarkable abilities to understand, reason, and generate texts across a variety of open-ended tasks. %
Recent studies further exploit the strong generality of LLMs to improve visual understanding or generation tasks, collectively referred to as Multimodal LLM (MLLM).
While these studies have contributed to technological advancements, MLLMs have yet to achieve the remarkable success of LLMs in terms of emergent capabilities.
We have made a bold assumption that the premise for the emergence of multimodal capabilities is that text and images can be represented and processed \textbf{interchangeably} in a unified autoregressive Transformer.

 \renewcommand{\thefootnote}{\fnsymbol{footnote}}
 		\footnotetext[1]{Equal Contribution. } 
   \footnotetext[2]{Correspondence to \texttt{yixiaoge@tencent.com}.}

We posit that a proper visual tokenizer is the key as it can facilitate the follow-up multimodal training by (i) easing the semantic alignment between visual and word tokens, and (ii) enabling LLM's original training recipe (i.e., next-word prediction) for multimodal data without specific adaptation for visual tokens.
Representing images as a sequence of discrete IDs is naturally compatible with the autoregressive training objective of LLMs.
But unfortunately, works \cite{ramesh2021zero,ding2021cogview} that utilize discretized visual tokens for multimodal tasks have receded from prominence, as such models generally rely on super-scale training to converge, leading to substantial training costs.
Moreover, our previous work~\cite{ge2023planting} empirically found that the dominant tokenizer VQ-VAE~\cite{van2017neural} in existing works captures too low-level information for LLMs to effectively perform multimodal comprehension tasks.
Existing image tokenizers fail to meet the requirements of unifying the generation of images and texts and facilitating multimodal training.

\begin{figure}[t]
\vspace{-10pt}
	\centering
	\includegraphics[width=1.0\linewidth]{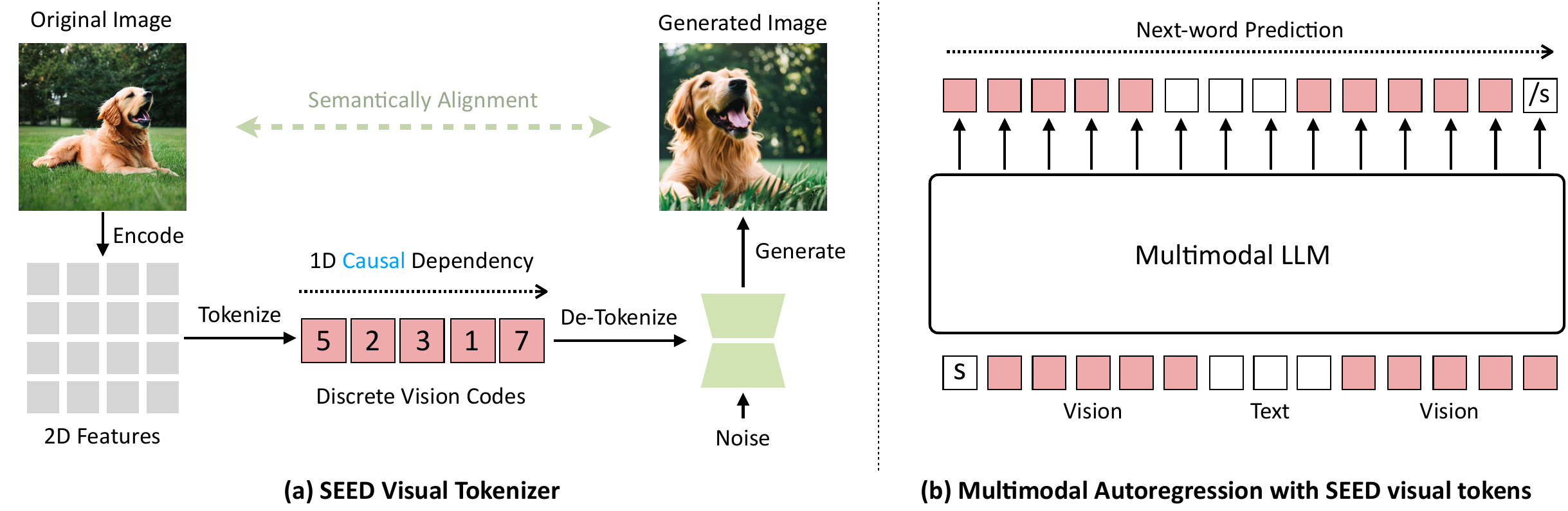}
\caption{(a) SEED is a discrete image tokenizer, producing quantized visual codes with 1D causal dependency and high-level semantics. (b) With SEED tokenizer, LLM is able to perform scalable multimodal autoregression on interleaved visual and textual data with next-word-prediction objective.}
\vspace{-10pt}
	\label{fig:teaser}
\end{figure}

To this end, we introduce \textbf{SEED}, a VQ-based image tokenizer that produces discrete visual codes with 1D causal dependency and necessary high-level semantics for both visual comprehension and generation tasks, as shown in Fig. \ref{fig:teaser} (a). 
The off-the-shelf LLMs can be readily equipped with SEED by treating discrete visual tokens as new words and updating the vocabulary.
We would like to emphasize the design principles of SEED.
(1) \textit{Why causal-dependent tokens?}
Existing visual tokens (\textit{e.g.}, from VQ-VAE or CLIP-ViT~\cite{sun2023eva}) are generated using 2D context, which is incompatible with the unidirectional attention in dominant LLMs and counterintuitive for text-to-image tasks requiring raster order prediction. Thus, we convert 2D raster-ordered embeddings into a sequence of semantic codes with 1D causal dependency.
(2) \textit{Why high-level semantics?}
Since visual and textual tokens in LLMs are expected to be interoperable—sharing weights and training objectives—they should encompass the same degree of semantics to prevent misalignment, i.e., the high-level semantics inherently present in words.

Specifically, the SEED tokenizer is composed of a ViT encoder~\cite{dosovitskiy2020image}, Causal Q-Former, VQ Codebook~\cite{van2017neural}, multi-layer perceptron (MLP), and a UNet decoder~\cite{ronneberger2015u}.
The ViT encoder and UNet decoder are directly derived from the pre-trained BLIP-2~\cite{li2023blip} and unCLIP-SD model~\cite{rombach2022high,ramesh2022hierarchical}, respectively.
(1) \textit{Tokenize:}
Causal Q-Former converts 2D raster-ordered features produced by the ViT encoder into a sequence of causal semantic embeddings, which are further discretized by the VQ Codebook.
(2) \textit{De-Tokenize:}
The discrete visual codes are decoded into generation embedding via MLP. The generation embedding is aligned with the latent space of unCLIP-SD so that realistic images with consistent semantics can be generated using the off-the-shelf SD-UNet.

We further present \textbf{SEED-LLaMA} by equipping the pre-trained LLM~\cite{touvron2023llama} with SEED tokenizer.
SEED-LLaMA is pretrained on multimodal data, including image-text pairs, video-text pairs, and interleaved image-text data, toward the training objective of next-word prediction as shown in Fig. \ref{fig:teaser} (b).
Such an easy-to-implement and unified proxy task facilitates scalable multimodal pretraining.
We further apply multimodal instruction tuning to align SEED-LLaMA with human instructions through supervised fine-tuning. 
Our model demonstrates extensive emergent abilities such as multi-turn in-context image and text generation given multimodal instructions as shown in Fig.~\ref{fig:teaser_example}. We also benchmark on a broad range of tasks including image captioning, image/video question answering, and text-to-image generation, receiving competitive performance.

In summary, our contributions are three-fold. 
(1) We introduce SEED, an advanced image tokenizer, designed based on the insights that visual tokens compatible with LLMs should capture high-level semantics while being generated with 1D causal dependency. The tailored SEED improves the scalability of subsequent multimodal training.
(2) We present SEED-LLaMA, composed of a pretrained LLM and SEED tokenizer, through large-scale multimodal pretraining and instruction tuning under the next-word-prediction training objective. It successfully unified multimodal comprehension and generation tasks in one framework.
(3) SEED-LLaMA shows competitive results on existing multimodal tasks ({e.g.}, text-to-image, image-to-text) and further demonstrates emergent abilities in multi-turn in-context multimodal understanding, reasoning, and generation.

\section{Related Work}
{\flushleft \bf MLLMs for Comprehension and Generation.} 
With the impressive success of Large language models~\cite{touvron2023llama, brown2020language, chowdhery2022palm} (LLMs), recent studies work on Multimodal LLM (MLLM) to improve visual \textbf{comprehension} through utilizing the strong generality of LLMs. Previous work~\cite{ye2023mplug,li2023blip,zhu2023minigpt,zhang2023llama,gao2023llama,liu2023visual,alayrac2022flamingo,driess2023palm} align visual features of pre-trained image encoder with LLMs on image-text datasets. However, these work commonly use the prediction of the next \textit{text token} as the objective, thus can only output texts.

To empower LLMs with the image \textbf{generation} ability, CogView~\cite{ding2021cogview} pre-trains a visual tokenizer by reconstructing image pixels, and fine-tunes GPT ~\cite{brown2020language} with the objective of next-token prediction. GILL~\cite{koh2023generating} learns a mapping between the embeddings of a LLM and a frozen text-to-image generation model. Both work aim to generate images with LLMs, without being explicitly designed for unifying multimodal comprehension and generation. 

Our concurrent works~\cite{{sun2023generative,yu2023scaling}} both perform multimodal autoregression including the generation of images and texts. CM3Leon~\cite{yu2023scaling} utilizes discrete visual codes from a image tokenizer~\cite{gafni2022make} pre-trained on image pixel reconstruction and performs image-to-text and text-to-image autoregression. However, it yields suboptimal performance in visual comprehension tasks ({e.g.}, CIDEr 61.6 vs. ours 126.9 on COCO image captioning) because the image tokenizer captures too low-level information. Emu~\cite{sun2023generative} employs continuous visual representations and is pre-trained on interleaved multimodal sequences through classifying the next text token or \textbf{regressing} the next visual embedding. For image generation, Emu further fine-tunes a SD model to accommodate the output representations from the LLM. By contrast, we pre-train a discrete image tokenizer, where the visual codes can be decoded to realistic images using the off-the-shelf SD model, and perform multimodal autoregressive with a unified next-word-prediction objective, which facilitates scalable multimodal training.

{\flushleft \bf Visual Tokenizer.} Visual tokenizer aims to represent images as a sequence of discrete tokens. Previous work~\cite{van2017neural,ramesh2021zero,esser2021taming,gu2022rethinking} trains a Vector Quantized Variational AutoEncoders (VQ-VAE) by reconstructing image pixels, which captures only low-level details such as color, texture and edge. Beit v2~\cite{peng2022beit} trains a visual tokenizer through reconstructing high-level features from the teacher model, but its visual codes from 2D features of a vision transformer~\cite{dosovitskiy2020image} are incompatible with the unidirectional attention in dominant LLMs for image generation. By contrast, we present SEED tokenizer, which produces discrete visual codes with 1D causal dependency and high-level semantics.

\section{Method}

\begin{figure}
	\centering
	\includegraphics[width=1.0\linewidth]{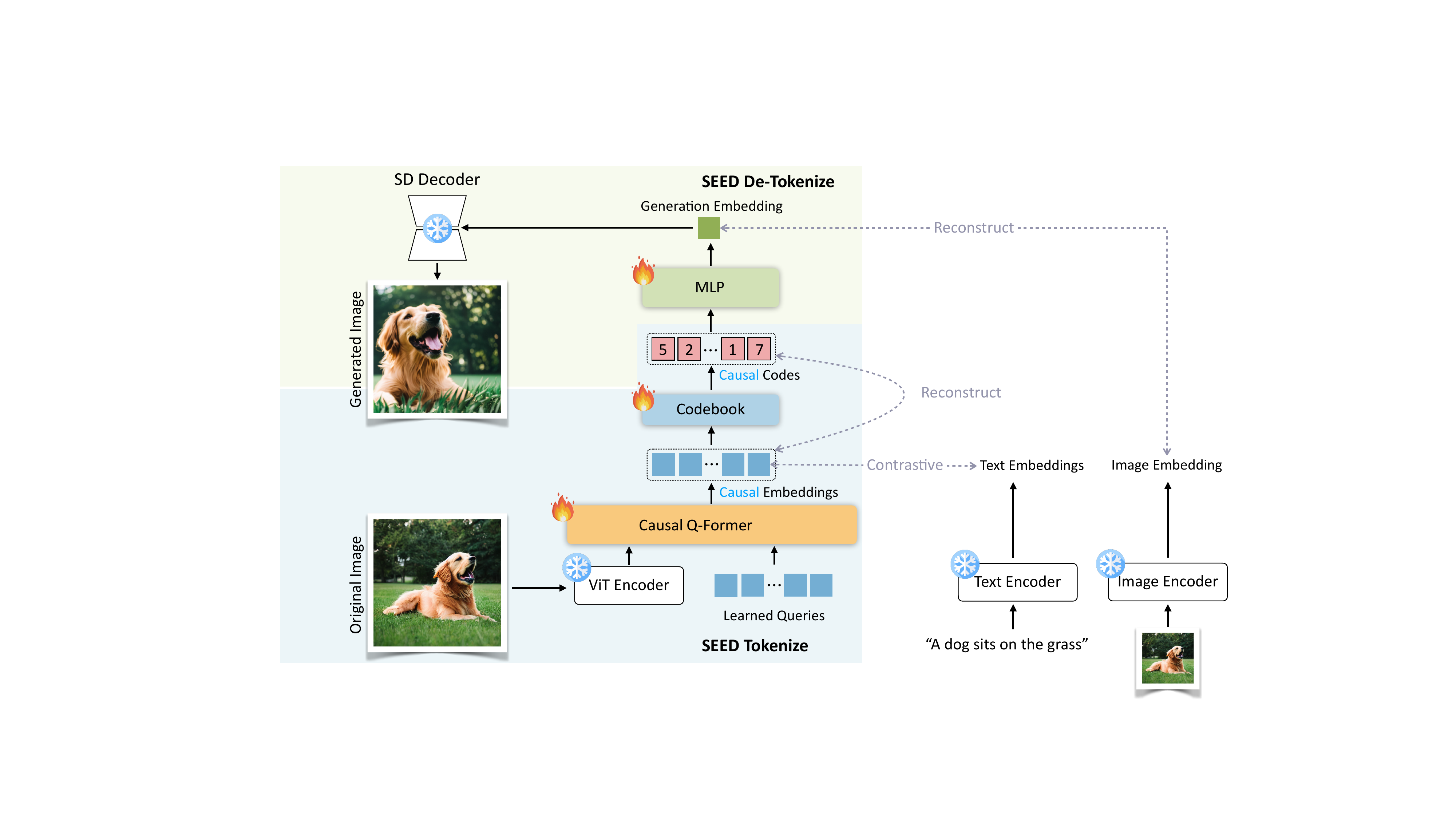}
  \vspace{-20pt}
\caption{Overview of \textbf{SEED} tokenizer, which produces discrete visual codes with causal dependency and high-level semantics. The generation embedding from visual codes can be decoded to realistic images with the frozen unCLIP~\cite{ramesh2022hierarchical} SD, which is conditioned on image embedding.}
 \vspace{-10pt}
	\label{fig:tokenizer}
\end{figure}

\subsection{SEED Tokenizer}\label{sec:tokenizer} 
As shown in Fig.~\ref{fig:tokenizer}, the SEED tokenizer is composed of  a ViT encoder~\cite{dosovitskiy2020image}, Causal Q-Former, VQ Codebook~\cite{van2017neural}, multi-layer perceptron (MLP), and a UNet decoder~\cite{ronneberger2015u}.
The ViT encoder and UNet decoder are directly derived from the pre-trained BLIP-2~\cite{li2023blip} and unCLIP~\cite{ramesh2022hierarchical} Stable Diffusion (unCLIP-SD) ~\cite{rombach2022high}, respectively. We first train a Causal Q-Former to convert 2D raster-ordered features (16$\times$16 tokens) produced by the ViT encoder into a sequence of causal embeddings (32 tokens). We then train a visual codebook to discretize the causal embeddings to quantized visual codes (32 tokens) with causal dependency. We employ a MLP to decode the visual codes into generation embedding (1 token), which is aligned with the latent space of the pre-trained unCLIP-SD conditioned on image embedding. Our previous work~\cite{ge2023planting} aligns generation embeddings with the text embeddings of SD~\cite{rombach2022high}, and we analyze the difference in Sec.~\ref{sec:ablation}. We pre-train SEED tokenizer on CC3M \cite{sharma2018conceptual}, Unsplash \cite{Unsplash}, LAION-COCO \cite{laion-coco} and MS-COCO \cite{chen2015microsoft}. 

\subsubsection{Training Stage I: Causal Q-Former}
As shown in Fig.~\ref{fig:tokenizer}, a set number of learnable query embeddings (32 tokens) and features of a pre-trained ViT encoder~\cite{sun2023eva} are fed into the Causal Q-former to encode a fixed number of causal embeddings (32 tokens) of the input image. Specifically, the query embeddings can interact with only previous queries through self-attention layers with causal mask, and interact with frozen image features through cross-attention layers. We adopt contrastive learning to optimize Causal Q-former fine-tuned from BLIP-2 Q-Former on 
image-text pairs. We use contrastive loss to maximize the similarity between the \textbf{final} causal embedding and text features of the corresponding caption.

\subsubsection{Training Stage II: Visual Tokenize and De-tokenize}\label{sec:vq} 

As shown in Fig.~\ref{fig:tokenizer}, we train a VQ codebook to discretize the causal embeddings (32 tokens) into quantized visual codes (32 tokens). Specifically, a quantizer looks up the nearest neighbor in the codebook for each causal embedding and obtains the corresponding code. We employ a decoder, which is a multi-layer Transformer~\cite{dosovitskiy2020image}, to reconstruct the continuous causal embeddings from discrete codes. During training, we maximize the cosine similarity between the output of the decoder and the causal embeddings. We further employ a MLP to reconstruct the image embedding (1 token) of a frozen unCLIP-SD from discrete codes. During training, we minimize the MSE loss between the generation embedding and the image embedding of unCLIP-SD. During inference, the generation embedding are fed into the off-the-shelf SD-UNet to decode realistic images.

\subsection{SEED-LLaMA}

\subsubsection{Training Stage I: Multimodal Pretraining}\label{sec:pretrain} 
\begin{figure}
	\centering
	\includegraphics[width=1.0\linewidth]{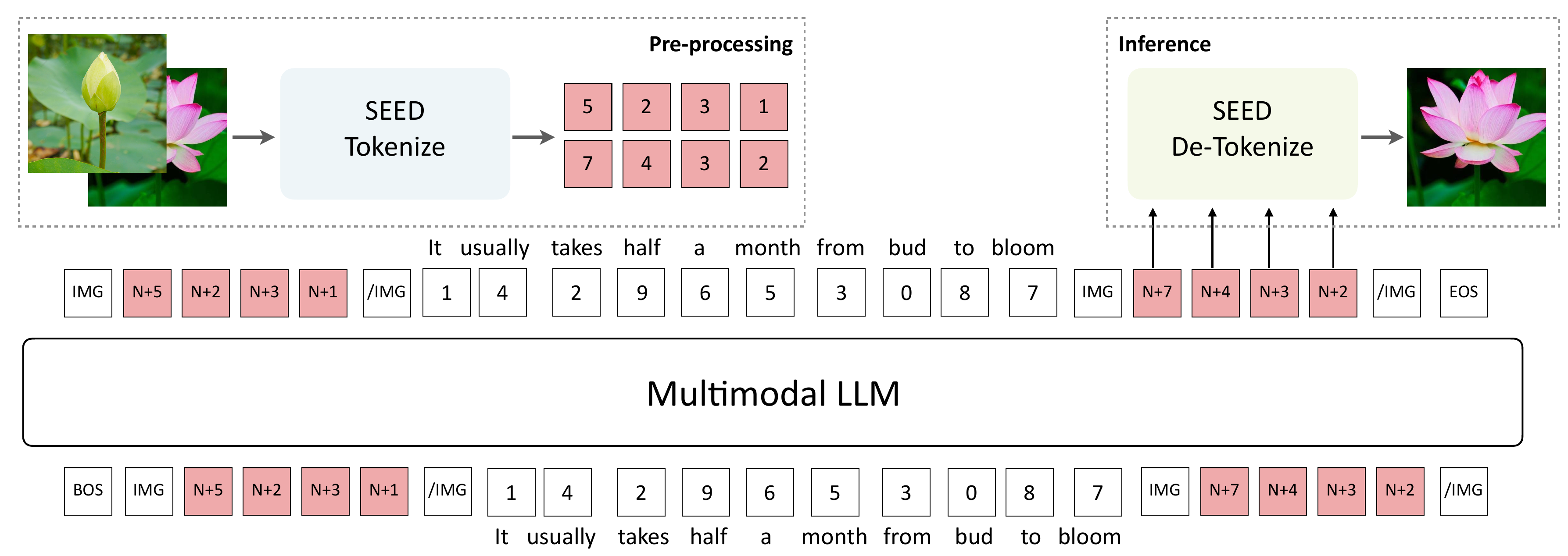}
\caption{Overview of the multimodal autoregressive pretraining on interleaved visual and textual data for \textbf{SEED-LLaMA}. Visual inputs are pre-processed into discrete tokens to conserve computational resources. Given the multimodal discrete sequence, a unified next-word-prediction objective is employed. During inference, visual codes are decoded into a realistic image by SEED De-Tokenization.}
	\label{fig:framework}
 \vspace{-10pt}
\end{figure}

As shown in Fig.~\ref{fig:framework}, SEED-LLaMA adopts a unified next-word-prediction training objective on interleaved visual and textual data. Specifically, visual inputs are first discretized into a sequence of causal codes by SEED tokenizer. Then the interleaved visual codes and text tokens are fed into the pretrained LLM for performing multimodal autoregression, where the visual codes are treated as new words and the vocabulary of the LLM is updated accordingly. We maximize the likelihood in a unified autoregressive manner as follows:
\begin{equation}
L(\mathcal{U})=\sum_i \log P\left(u_i \mid u_{i-k}, \ldots, u_{i-1} ; \Theta\right)
\label{func:autoregression}
\end{equation}
where $u_i$ represents visual code or text token, and $\Theta$ denotes the the parameters of the transformer. We initialize SEED-LLaMA from a pre-trained LLM, and add 8192 visual codes to the vocabulary. The embedding layer and decoder head layer in the transformer are expanded and the parameters of added visual codes are randomly initialized. 

For efficiency, we first train SEED-LLaMA using LoRA~\cite{hu2021lora} tuning and together optimize the parameters of the embedding layer and decoder head layer due to the added visual codes. We then merge the parameters of LoRA onto the LLM backbone and fine-tune all parameters except for the embedding layer. We freeze the embedding layer since we observe that fine-tuning it together with other parameters can lead to unstable training loss, which is also reported in BLOOM \cite{scao2022bloom} and GLM-130B \cite{zeng2022glm}. We preprocess the images and videos into discrete tokens beforehand to conserve computational resources. We perform pretraining using two versions of LLM, Vicuna-7B and Llama2-chat-13B, with 64 A100-40G GPUs, and yield SEED-LLaMA-8B (144 hours) and SEED-LLaMA-14B (216 hours), respectively. See Appendix.~\ref{sec:pretrain_appendix} for details.

\subsubsection{Training Stage II: Multimodal Instruction Tuning}\label{sec:sft} 
We perform multimodal instruction tuning on SEED-LLaMA to align it with human instructions through supervised finetuning on public datasets. The details of datasets can be found in Appendix.~\ref{sec:appendix_sft}. We fine-tune a LoRA module on the pre-trained SEED-LLaMA with the template as below,
\begin{equation}
\text{USER: \quad<Instruction>\quad ASSISTANT:\quad <Answer>} 
\label{func:instruction}
\end{equation}
Only the content of <Answer> is accounted for loss. The overall instruction tuning phase takes 16 hours for SEED-LLaMA-8B and 27 hours for SEED-LLaMA-14B with 32 A100-80G GPUs.

\section{Experiment}

\subsection{SEED Tokenizer}\label{sec:exp_tokenizer}
{\flushleft \bf Evaluation of Causal Embeddings.} We evaluate the performance of Causal Q-Former on the image-text retrieval using COCO~\cite{lin2014microsoft} and Flickr30K~\cite{young2014image}. The performance is measured by \emph{Recall@K} (R@K). Note that we adopt the dual-stream paradigm for inference and remove the image-text-matching (ITM) re-rank module in BLIP-2 for a fair comparison. As shown in Tab.~\ref{tab:retrieval}, our Causal Q-former achieves better results than BLIP-2 in terms of an aggregated metric \emph{Recall@mean}. It demonstrates that the output query embeddings with causal dependency do not drop performance than the output embeddings with bi-directional attention in BLIP-2. 

\begin{table}[t]
\centering
   \vspace{-25pt}
\renewcommand\arraystretch{1.4}
\caption{Evaluation of Image-Text Retrieval. Causal codes are quantized causal embeddings.}
\resizebox{1.\columnwidth}{!}{
\begin{tabular}{lccccccc|ccccccc}
\toprule
\multirow{3}{*}{Model} & \multicolumn{7}{c|}{Flickr30K (1K test set)}                 & \multicolumn{7}{c}{COCO (5K test set)}                      \\
                                                      & \multicolumn{3}{c}{Image $\rightarrow$ Text} & \multicolumn{3}{c}{Text $\rightarrow$ Image} & &\multicolumn{3}{c}{Image $\rightarrow$ Text} & \multicolumn{3}{c}{Text $\rightarrow$ Image}& \\
\cmidrule(l){2-15}
                                                         & R@1        & R@5      & R@10      & R@1       & R@5       & R@10   &R@m   & R@1       & R@5       & R@10      & R@1       & R@5       & R@10 &R@m     \\
\midrule
BLIP-2 ~\cite{li2023blip}                                     & 81.9       & 98.4     & 99.7      & \textbf{82.4}      & \textbf{96.5}      & \textbf{98.4}     &92.9 & 65.3      & 89.9      & 95.3      & \textbf{59.1}      & 82.7      & \textbf{89.4}  &80.3   \\
SEED (causal embedding) &\textbf{91.0}	&\textbf{99.5}	&\textbf{100.0}	&79.3	&94.8	&97.1	&\textbf{93.6}	&\textbf{74.2}	&\textbf{93.1}	&\textbf{96.7}	&59.0	&\textbf{82.8}	&89.2	&\textbf{82.5}\\
SEED (causal code) &85.4	&98.3	&99.6	&73.7	&92.3	&95.7	&90.8	&66.9	&89.3	&94.4	&53.2	&78.8	&86.6	&78.2\\
\bottomrule
\label{tab:retrieval}
\vspace{-15pt}
\end{tabular}}
\end{table}

\begin{figure}
	\centering
	\includegraphics[width=1.0\linewidth]{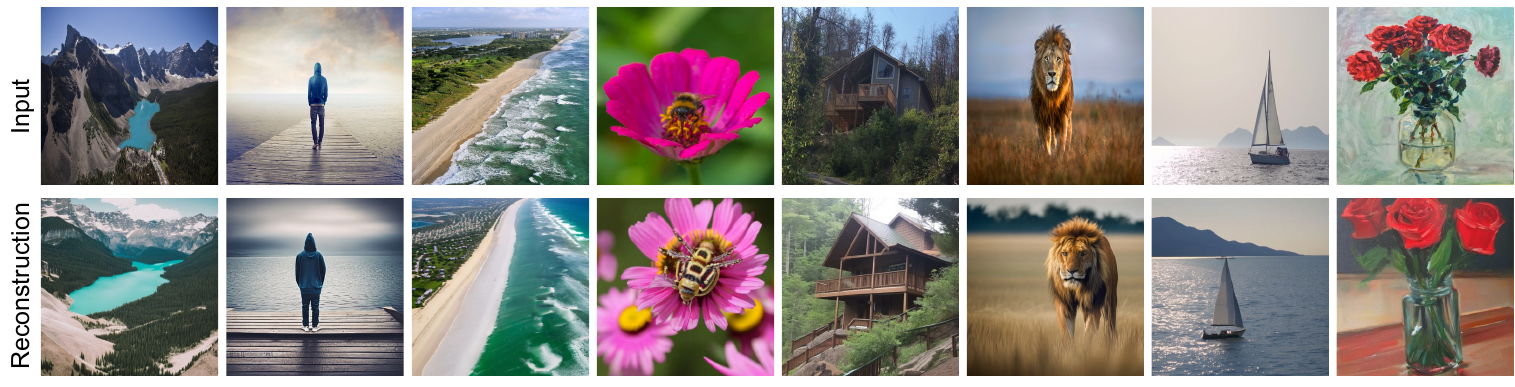}
 \vspace{-15pt}
\caption{Reconstruction images of SEED tokenizer (\textit{i.e.}, original image $\rightarrow$ SEED tokenize $\rightarrow$ causal visual codes $\rightarrow$ SEED de-tokenize $\rightarrow$ reconstructed image).}
	\label{fig:recon}
  \vspace{-10pt}
\end{figure}

{\flushleft \bf Evaluation of Causal Codes.} We evaluate causal codes on the image-text retrieval, where the reconstructed embeddings from causal codes are used for retrieval.  As shown in Tab.~\ref{tab:retrieval}, discrete codes exhibit competitive performance compared to BLIP-2, which demonstrates that the discrete codes from SEED tokenizer capture high-level semantics, which are suitable for visual comprehension. 

\begin{wraptable}{r}{6cm}
\centering
\vspace{-20pt}
\caption{Evaluation of Image Generation. }
\resizebox{.4\columnwidth}{!}{
\begin{tabular}{lcc}
\toprule
Model & COCO  & Flickr30K \\
\midrule
\textit{Image-to-image} &\\
unCLIP~\cite{ramesh2022hierarchical} SD &\textbf{79.30} &\textbf{79.55} \\
$\text{SEED}^{\text{text}}$~\cite{ge2023planting}    & 68.23 & 65.22      \\
SEED    & 77.35 & 76.52      \\
\hline
\textit{Text-to-image} &\\
GILL~\cite{koh2023gill}  & 67.45 & 65.16     \\
Emu~\cite{sun2023generative} &66.46 &64.82 \\
SEED-LLaMA    &69.07  &65.54       \\
SEED-LLaMA-I &\textbf{70.68} &\textbf{66.55}\\
\bottomrule
\label{tab:clip_score}
\vspace{-25pt}
\end{tabular}}
\end{wraptable}
We further evaluate image reconstruction on COCO and Flickr30K dataset. SEED first discretizes input images into causal codes (32 tokens) and obtain generation embedding (1 token), which are fed into the unCLIP-SD-UNet for reconstruction. We follow GILL~\cite{koh2023generating} to compute the CLIP similarity score as the metric to evaluate the semantic consistency. 
As shown in Tab.~\ref{tab:clip_score}, compared with the upper bound unCLIP-SD, SEED only slightly drops performance.

We visualize the reconstructed images of SEED tokenizer in Fig.~\ref{fig:recon}. Through obtaining the generation embedding from the causal visual codes, realistic images can be generated using the frozen SD-UNet, which maintain consistent semantics with inputs. 
\textit{The above evaluation and visualization demonstrate the versatility of SEED visual tokens for both comprehension and generation tasks.}

\subsection{SEED-LLaMA}\label{sec:SEED-LLaMA}
\begin{table}
\centering
\small
\setlength\tabcolsep{3pt}
\caption{Comparison for multimodal comprehension. ``Image Gen'' denotes whether the model can generate images besides texts, and ``-I'' denotes the instruction tuned model. The best results are \textbf{bold} and the second best are \uline{underlined}.}
\label{tab:sticker_clip_eval}
\resizebox{0.96\columnwidth}{!}{
\begin{tabular}{l|c|c|ccccc|ccc} 
\toprule
\multirow{2}{*}{Models} &\multirow{2}{*}{Size}&\multirow{2}{*}{Image}& \multicolumn{5}{c|}{Image-Text Tasks    }                                                                             & \multicolumn{3}{c}{Video-Text Tasks   }                                                               \\
                        &&Gen& COCO           & \small{VQAv2}         & \small{OKVQA}         & VizWiz        & \begin{tabular}[c]{@{}c@{}}\small{SEED} \\Bench\end{tabular} & \small{MSVDQA}         & \small{MSRVTTQA}      & \small{NExTQA}          \\ 
\hline
Flamingo~\cite{alayrac2022flamingo}&9B&$\times$&79.4 &51.8 &44.7&28.8&42.7&30.2&13.7&23.0\\
BLIP-2~\cite{li2022blip} &4.1B &$\times$ &\textbf{144.5} &63.0 &40.7 &29.8 &49.7 &33.7 &16.2 &-\\
InstructBLIP~\cite{li2023blip} &8.1B&$\times$&-&-&-&34.5&\textbf{58.8}&\uline{41.8}&22.1&-\\
Kosmos-1~\cite{huang2023language}          &1.6B  &$\times$& 84.7           & 51.0          & -             & 29.2          & -                                                    & -              & -                        & -                                                    \\                                         
Kosmos-2~\cite{peng2023kosmos} &1.6B &$\times$&-&45.6&-&-&54.4&-&-&-\\
MetaLLM~\cite{hao2022language}          &1.7B  &$\times$ & 82.2           & 41.1          & 11.4          & -             & -                                                    & -              & -                        & -                                                    \\
IDEFICS~\cite{laurencon2023obelics}            &80B  &$\times$& 91.8           & 60.0          & \uline{45.2}  & 36.0          & -                                                    & -              & -             & -                                                                \\
IDEFICS-I~\cite{laurencon2023obelics}          &80B  &$\times$& 117.2          & 37.4          & 36.9          & 26.2          &53.2                                                    & -              & -             & -                                                               \\
CM3Leon~\cite{yu2023scaling}         &7B   &$\checkmark$& 61.6           & 47.6          & 23.8          & 37.6          & -                                                    & -              & -             & -                                                               \\
Emu~\cite{sun2023generative}                 &14B  &$\checkmark$& 112.4          & 52.0          & 38.2          & 34.2          & 47.3                                                 & 18.8           & 8.3           & 19.6                                                          \\
Emu-I~\cite{sun2023generative}              &14B &$\times$& 117.7          & 40.0          & 34.7          & 35.4          & \uline{58.0}                                        & 32.4           & 14.0          & 6.8                                                \\ 
\hline                                                 
\textbf{SEED-LLaMA}  &8B& $\checkmark$&123.6          & 44.2          & 29.2          & 21.5          & 42.2                                                 & 11.5           & 5.0           & 14.3             \\
\textbf{SEED-LLaMA-I}     &8B&$\checkmark$&  124.5          & \textbf{66.2} & \textbf{45.9} & \textbf{55.1} & 51.5                                                 & 40.9   & \uline{30.8}  & \textbf{24.9}                                               \\                                             
\textbf{SEED-LLaMA} &14B&$\checkmark$& 125.0  & 48.1          & 27.1          & 23.3          & 46.0                                                 & 13.9           & 3.7           & 11.3                                                \\
\textbf{SEED-LLaMA-I}    &14B&$\checkmark$&  \uline{126.9} & \uline{63.4}  & 43.2          & \uline{49.4}  & 53.7                                         & \textbf{45.2} & \textbf{35.3} & \uline{24.7}\\
\bottomrule
\end{tabular}}
\end{table}

\begin{figure}
	\centering
	\includegraphics[width=1.0\linewidth]{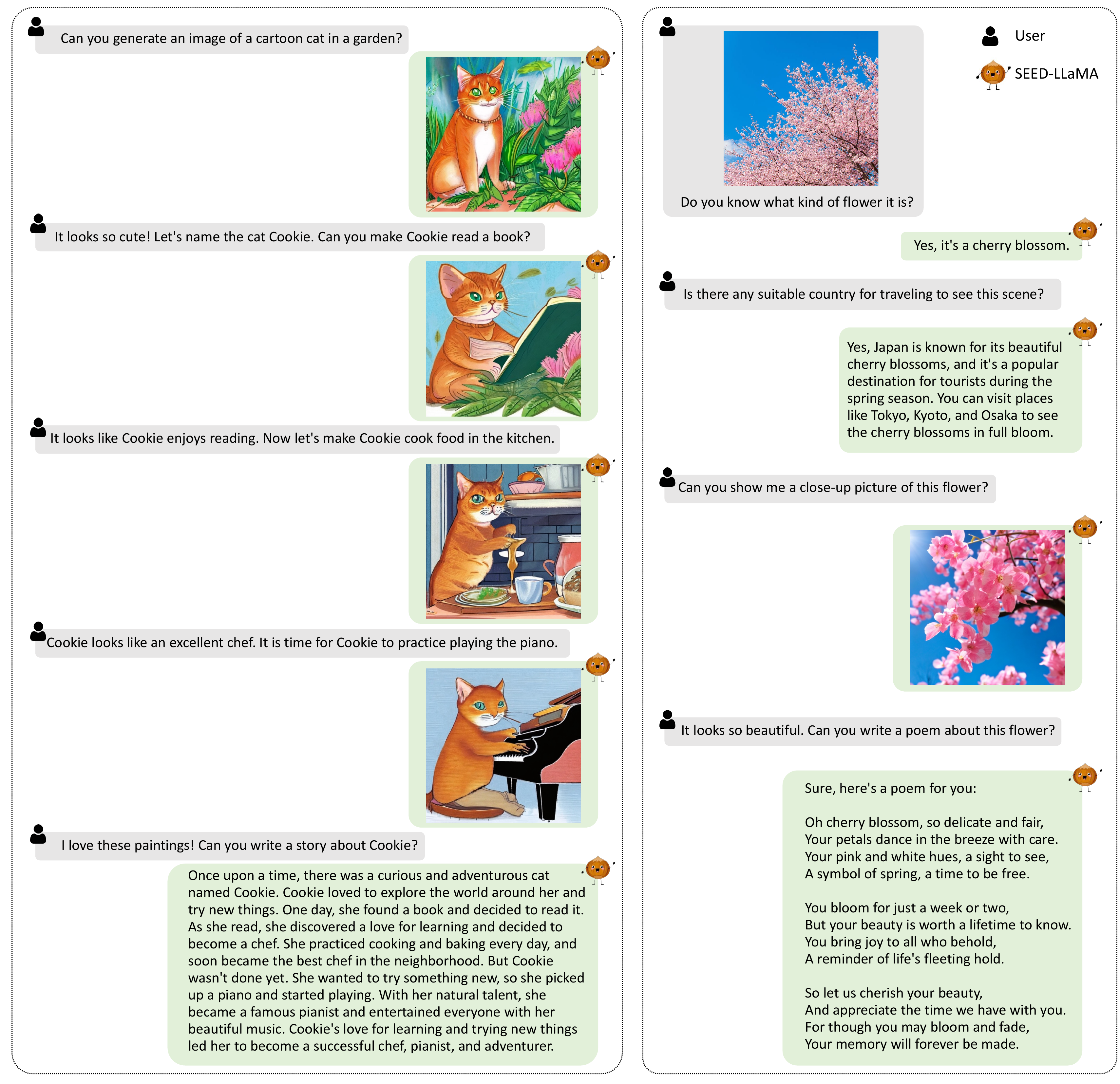}
\caption{Qualitative examples of multi-turn in-context image and text
generation by SEED-LLaMA given multimodal instructions.}
	\label{fig:multi_turn}
  \vspace{-10pt}
\end{figure}

\subsubsection{Quantitative Evaluation}
{\flushleft \bf Multimodal Comprehension.} We evaluate SEED-LLaMA on a wide range of multimodal comprehension tasks including image captioning and image/video question answering. Details of these benchmarks and evaluation metrics are provided in Appendix.~\ref{sec:evaluation}. As shown in Tab.~\ref{tab:sticker_clip_eval}, our SEED-LLaMA achieves competitive performance in both the image and video understanding tasks compared with MLLMs that use continuous visual representations. The results demonstrate that our SEED tokenizer can generate discrete visual codes with high-level semantics, which facilities the visual comprehension. We can observe that pretraining from a LLM with larger model size improves performance on SEED-Bench and instruction tuning further contributes to enhanced results. Note that as pointed out by recent work~\cite{liu2023mmbench,li2023seed}, previous VQA benchmarks listed in Tab.~\ref{tab:sticker_clip_eval} are not tailored for evaluating MLLMs with open-from output, since they require an exact match between the model prediction and the target word or phrase. 
The qualitative examples of multimodal comprehension is provided in Appendix.~\ref{sec:qualitative}.

{\flushleft \bf Text-to-image Generation.} 
We evaluate the text-to-image generation on MS-COCO~\cite{chen2015microsoft} and Flickr30K~\cite{young2014image} and compute the pair-wise CLIP similarity score as the evaluation metric following GILL~\cite{koh2023gill}. As shown in Tab.~\ref{tab:clip_score}, images generated by our SEED-LLaMA from textual descriptions show higher similarity with the ground-truth images. The results demonstrate that SEED-LLaMA generates images that are highly correlated with text prompts via a frozen SD-UNet. We show qualitative examples of text-to-image generation in Appendix.~\ref{sec:qualitative}.

\begin{figure}
	\centering
  \vspace{-15pt}
	\includegraphics[width=1.0\linewidth]{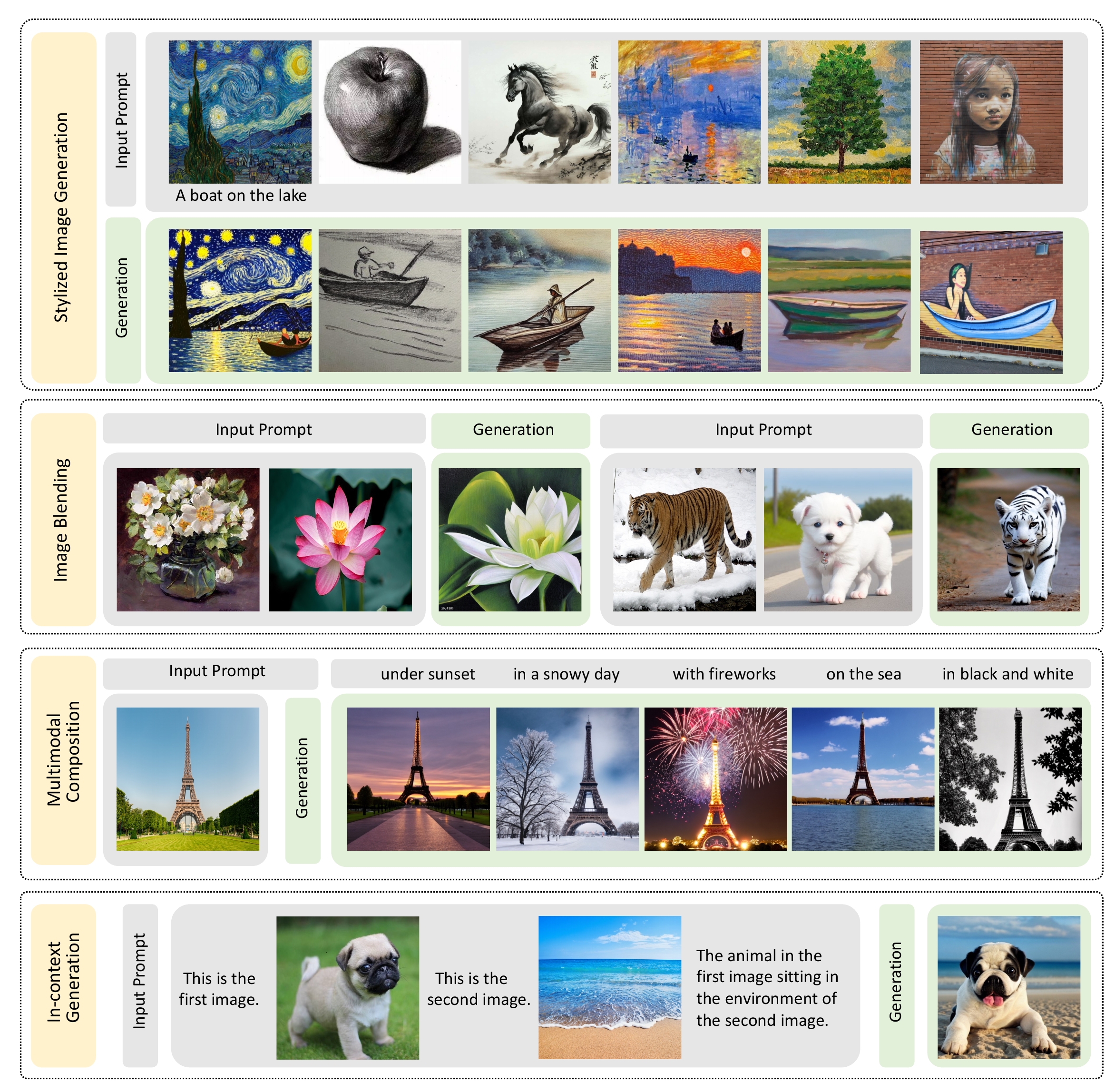}
\caption{Qualitative examples of compositional image generation by SEED-LLaMA.}
	\label{fig:composition}
 \vspace{-10pt}
\end{figure}

\subsubsection{Emergent Ability}
{\flushleft \bf Multi-turn In-context Multimodal Generation.}
As shown in Fig.~\ref{fig:teaser_example} and Fig.~\ref{fig:multi_turn}, given multimodal instructions including images and open-form texts from a user, our SEED-LLaMA can respond with synthesized image (\textit{e.g.}, a dog in front of the Golden Gate Bridge), 
sequentially generated images (\textit{e.g.}, a cartoon cat in different scenes), instruction-followed image (\textit{e.g.}, a closer look-up of a cherry blossom), various forms of texts via creation and real-world knowledge (\textit{e.g.}, a story, a poem and flower identification). The results illustrate the impressive capability of SEED-LLaMA in reasoning and generating long-context multimodal content.

{\flushleft \bf Compositional Image Generation.}
As shown in Fig.~\ref{fig:composition}, our SEED-LLaMA can realize a variety of zero-shot compositional image generation as below,
\begin{itemize}

\item Stylized Image Generation. SEED-LLaMA can take a text prompt and a style reference image as inputs and produce an output image that adheres to both the style and text prompt.

\item Image Blending. SEED-LLaMA can take two images as inputs and generate an image that blends the visual components of the input images.

\item Multimodal Composition. SEED-LLaMA can take an image prompt and a text prompt as inputs and generate a composite image that combines the multimodal inputs.

\item In-context Generation. SEED-LLaMA can take images, their textual references, and text prompts as inputs and generate context-related images.

\end{itemize}

\subsection{Ablation Study} \label{sec:ablation}

{\flushleft \bf Generation Embedding.} The generation embedding of SEED is aligned with the image embedding of unCLIP-SD, and can be decoded to realistic images with the unCLIP-SD-UNet. In our previous work~\cite{ge2023planting}, we train a visual tokenizer $\text{SEED}^{\text{text}}$ through aligning the generation embeddings with the text embeddings (77 tokens) of SD~\cite{rombach2022high} conditioned on texts. As shown in Tab.~\ref{tab:clip_score}, the similarity between the reconstructed images of $\text{SEED}^{\text{text}}$ and original images drop heavily. The semantic representations of texts can not fully preserve the rich visual information of images. The visual comparison of the the reconstructed images between $\text{SEED}^{\text{text}}$ and SEED are provided in Appendix.~\ref{sec:tokenizer_comparison}.  

{\flushleft \bf Causal Visual Codes vs. Bilateral Visual Codes.} We train a Causal Q-Former to convert 2D features produced by the ViT encoder into a sequence of causal semantic
embeddings, which are further discretized as causal visual codes. To verify whether the causal visual codes are necessary for compatibility with LLM, we train a visual tokenizer $\text{SEED}^{\text{Bi}}$, which produces bilateral visual codes from a pre-trained Q-Former with bilateral self-attention. We then pre-train $\text{SEED}^{\text{Bi}}$-$\text{LLM}^{\ast}$ and $\text{SEED}$-$\text{LLM}^{\ast}$ on image-text pairs and evaluate the text-to-image generation on COCO test set. Given 5000 captions of COCO, $\text{SEED}^{\text{Bi}}$-LLM only generates 2134 images successfully while $\text{SEED}$-$\text{LLM}^{\ast}$ generates 4997 images (Failure cases occur when the model predicts a number of visual tokens not equal to 32). The results demonstrate that the non-causal codes lead to highly unstable model performance since they contradict with the left-to-right autoregressive mechanism of LLM.

{\flushleft \bf SEED-LLaMA Pretraining.} We first train SEED-LLaMA using LoRA tuning, and then merge the parameters of LoRA with the original LLM and fine-tune all parameters except for the embedding layer. 
\begin{wraptable}{r}{6.2cm}
\centering
\vspace{-15pt}
\caption{Evaluation of image captioning and text-to-image generation on COCO test set.}
\resizebox{.4\columnwidth}{!}{
\begin{tabular}{ccc}
\toprule
Pretraining &Captioning  & Generation \\
\midrule
LoRA&124.5 &68.87 \\
LoRA + Fully&\textbf{125.0} &\textbf{69.07}\\
\bottomrule
\label{tab:ablation}
\vspace{-15pt}
\end{tabular}}
\end{wraptable}
To explore whether fully fine-tuning helps, we evaluate the performance of the model before and after fully fine-tuning on image captioning and text-to-image generation, with evaluation metric CIDEr and clip similarity score. Tab.~\ref{tab:ablation} shows that fully fine-tuning the LoRA tuned model enhances model's capability for both image comprehension and generation.

\section{Conclusion}
We present SEED, a discrete image tokenizer, designed based on the premise that visual tokens compatible with LLMs should capture high-level semantics while being generated with 1D causal dependency. SEED enables LLMs to be trained with multimodal data following the original recipe of text (\textit{i.e.}, next-word prediction), which is mature and scalable. We further present
SEED-LLaMA via multimodal pretraining and instruction tuning on the interleaved visual and textual data with SEED tokenizer. SEED-LLaMA not only exhibits remarkable performance across multimodal comprehension and image generation tasks, but also demonstrates extensive compositional emergent abilities. We hope that SEED would draw increased attention to visual tokenizers. A more rational visual tokenizer could substantially reduce the complexity of multimodal LLM training.

{\small
\bibliographystyle{unsrt}
\bibliography{SEED}
}

\clearpage
\appendix

\section{SEED Tokenizer}\label{sec:tokenizer_comparison}
\begin{figure}
	\centering
	\includegraphics[width=1.0\linewidth]{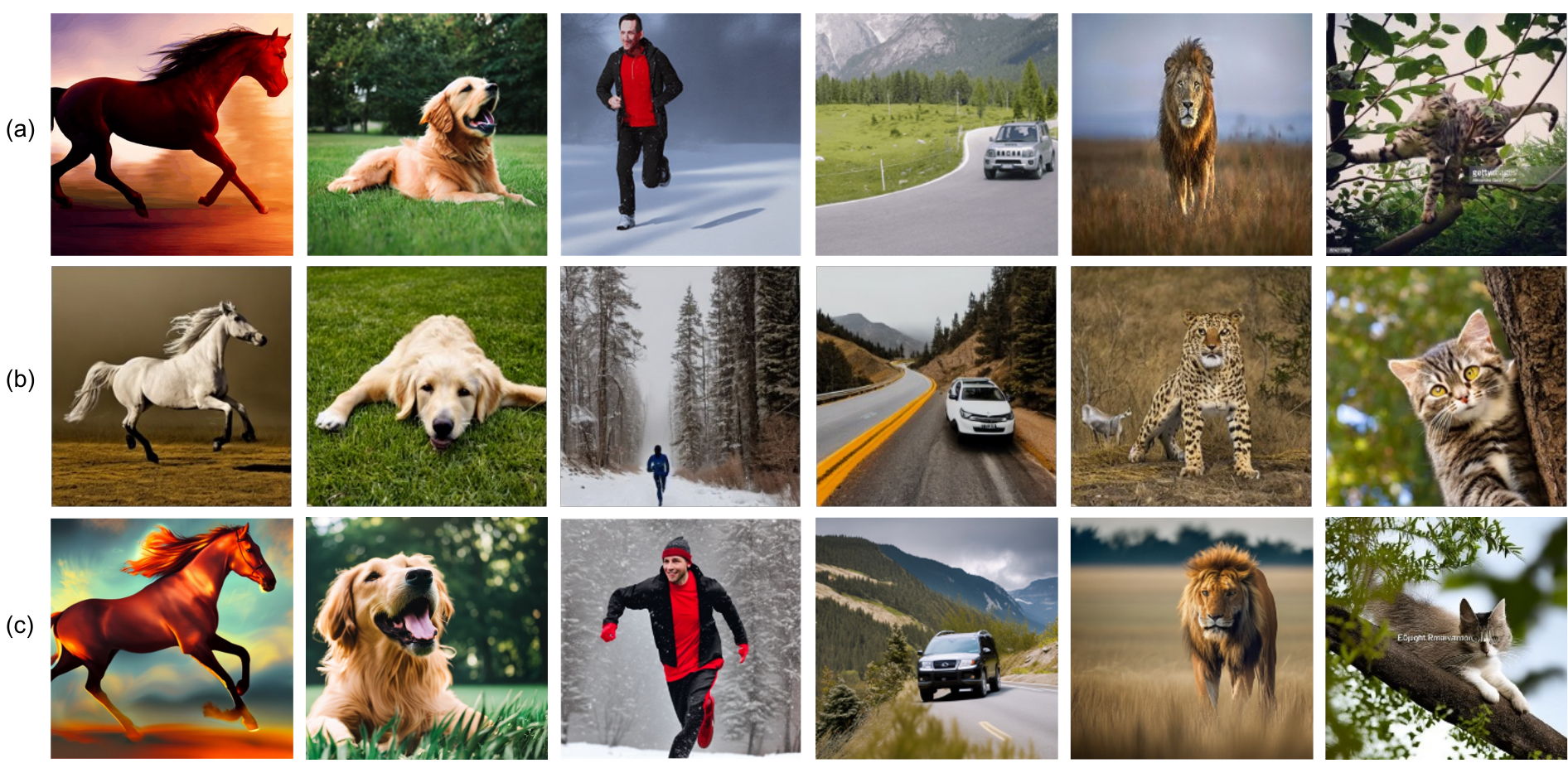}
\caption{(a) Input image. (b) Reconstruction images of $\text{SEED}^{\text{text}}$ tokenizer~\cite{ge2023planting}, which is aligned with the feature space of a SD conditioned on text embeddings. (c) Reconstruction images of SEED tokenizer, which is aligned with the feature space of a SD conditioned on image embedding.}
	\label{fig:seed_comparison}
 \vspace{5pt}
\end{figure}

The generation embedding of SEED is aligned with the image embedding of unCLIP~\cite{ramesh2022hierarchical} SD, and can be decoded to realistic images with the unCLIP-SD-UNet. In our previous work~\cite{ge2023planting}, we train a visual tokenizer $\text{SEED}^{\text{text}}$ through aligning the generation embeddings with the text embeddings (77 tokens) of SD~\cite{rombach2022high}, and the generation embeddings can be decoded to images with the SD-UNet. The visual comparison of the the reconstructed images between $\text{SEED}^{\text{text}}$ and SEED are shown in Fig.~\ref{fig:seed_comparison}. We can observe that compared with $\text{SEED}^{\text{text}}$, the images reconstructed by SEED can better preserve the visual information of the original images.

\section{Pretraining}\label{sec:pretrain_appendix}

\subsection{Pretraining Data}

As shown in Tab. \ref{tab:pretraing_data}, we utilize diverse categories of datasets as pretraining data, which can be summarized as follows.

{\flushleft \bf Image-text Pairs.}  We use the image-text pairs from CC3M \cite{sharma2018conceptual}, Unsplash \cite{Unsplash}, LAION-COCO \cite{laion-coco} and MS-COCO \cite{chen2015microsoft}. We filtered the samples in these datasets based on image resolution, aspect ratio, and visual-textual similarity. We randomly place images or text at the forefront, in order to achieve the generation of captions based on images and vice versa.

{\flushleft \bf Video-text Pairs.} We use a large-scale dataset WebVid-10M \cite{bain2021frozen} containing videos and captions. We implemented heuristic rules to exclude extraneous metadata, such as the resolution of the original video and camera parameters. We sample four frames of each video for training.

{\flushleft \bf Interleaved Image and Text.} We use publicly available MMC4 \cite{zhu2023multimodal} and OBELISC \cite{laurencon2023obelics} datasets, which were extracted and thoroughly filtered from Common Crawl. Specifically, we employ the MMC4-core split, consisting of 7.3 million samples, and the complete OBELISC dataset, containing 141 million samples. For documents in MMC4, we create a sequence of length 1024 and randomly shuffle the order of images and their corresponding texts (those with the highest CLIP score). As for OBELISC, we generate a sequence of length 1024 based on the order of data in the dataset.

\begin{table}
\centering
\renewcommand{\arraystretch}{1.1}
\caption{Description of pretraining datasets of SEED-LLaMA.} 
\label{tab:pretraing_data}
\begin{tabular}{ll} 
\toprule
Dataset Name & Dataset Description                                                                                                \\ 
\hline
\begin{tabular}[c]{@{}l@{}}COCO Caption\\ \cite{chen2015microsoft}\end{tabular} & \begin{tabular}[c]{@{}l@{}}0.5M image-text pairs with~human-written captions. Specifically,\\Karpathy train split is used.\end{tabular}                                                                                                                                                 \\ 
\hline
\begin{tabular}[c]{@{}l@{}}CC3M\\ \cite{sharma2018conceptual}\end{tabular}   & 3.3M image-text pairs~from the web.                                                                                                                                                                                                                                                  \\ 
\hline
\begin{tabular}[c]{@{}l@{}}Unsplash\\ \cite{Unsplash}\end{tabular}   & \begin{tabular}[c]{@{}l@{}}4.8M image-text pairs, in which images are composed of high-quality\\ Unsplash photos.\end{tabular}                                                                                                                                                            \\ 
\hline
\begin{tabular}[c]{@{}l@{}}LAION-COCO\\ \cite{laion-coco}\end{tabular}  & \begin{tabular}[c]{@{}l@{}}600M image-text pairs, where the caption is generated by the BLIP\\ \cite{li2022blip}.\end{tabular}                                                                                                                                                                                                                          \\ 
\hline
\begin{tabular}[c]{@{}l@{}}MMC4\\ \cite{zhu2023multimodal}\end{tabular}      & \begin{tabular}[c]{@{}l@{}}101M image-interleaved documents collected from Common Crawl.\\ We use the mmc4-core split which is consist of 7.3M documents. We\\randomly shuffle the order of images and their corresponding text\\(those with the~highest CLIP score).\end{tabular}  \\ 
\hline
\begin{tabular}[c]{@{}l@{}}OBELISC\\ \cite{laurencon2023obelics}\end{tabular}   & 141M image-interleaved documents collected from Common Crawl.                                                                                                                                                                                                                            \\ 
\hline
\begin{tabular}[c]{@{}l@{}}WebVid\\ \cite{bain2021frozen}\end{tabular}   & \begin{tabular}[c]{@{}l@{}}8M video-text pairs, we have implemented heuristic rules to exclude\\extraneous metadata,such as~the resolution of the original video and\\camera parameters.\end{tabular}                                                                                   \\
\bottomrule
\vspace{-20pt}
\end{tabular}
\end{table}

\subsection{Pretraining Hyperparameters}
We report the detailed pretraining hyperparameters of SEED-LLaMA in Tab. \ref{tab:pretraining_hyperparameters}. 

\begin{table}
\centering
\setlength\tabcolsep{9pt}
\renewcommand{\arraystretch}{1.2}
\caption{Summary of pretraining hyperparameters of SEED-LLaMA.}
\label{tab:pretraining_hyperparameters}
\begin{tabular}{lcc} 
\toprule
Configuration               & SEED 8B               & SEED 14B                                                                                            \\ 
\hline
Vision encoder              & \multicolumn{2}{c}{EVA-CLIP}                                                                                                \\
LLM                         & Vicuna-7B            & LLaMA2-Chat-13B                                                                                   \\
Training Strategy           & \multicolumn{2}{c}{LoRA + Fully fine-tuning}                                                                                \\
Peak learning rate          & \multicolumn{2}{c}{1.5e-4}                                                                                                  \\
Warmup ratio                & \multicolumn{2}{c}{0.03}                                                                                                    \\
LR schedule                 & \multicolumn{2}{c}{Cosine decay}                                                                                            \\
Optimizer                   & \multicolumn{2}{c}{AdamW}                                                                                                   \\
Optimizer hyper-parameters  & \multicolumn{2}{c}{$\beta_1$,$\beta_2$, $\epsilon$ = 0.9, 0.98, le-6}                                                       \\
Image resolution            & \multicolumn{2}{c}{224 $\times$ 224}                                                                                        \\
Weight decay                & \multicolumn{2}{c}{0.05}                                                                                                    \\
Iterations                  & \multicolumn{2}{c}{30k + 10k}                                                                                               \\
Data                        & \multicolumn{2}{c}{\begin{tabular}[c]{@{}c@{}}(MS-COCO, CC3M, Unsplash), LAION-COCO, \\OBELISC, MMC4, WebVid\end{tabular}}  \\
Sequence length per dataset & \multicolumn{2}{c}{160, 128, 1024, 1024, 200}                                                                               \\
Batch size per dataset      & 146, 180, 26, 26, 116 & 46, 56, 8, 8, 36                                                                                    \\
Sample ratio per dataset    & \multicolumn{2}{c}{4.5\%, 54.5\%, 9.1\%, 27.3\%, 4.5\%}                                                                     \\
\bottomrule
\end{tabular}
\end{table}

\section{Instruction Tuning}\label{sec:appendix_sft}
We summarize the datasets and their prompts for supervised instruction tuning of SEED-LLaMA in Tab.~\ref{tab:sft_data} and Tab.~\ref{tab:sft_prompt}. Note that MagicBrush~\cite{zhang2023magicbrush} contains both the single-turn and multi-turn scenarios, and we only use the single-turn for multimodal prompt image generation.

\begin{table}
\small
\centering
\caption{Description of datasets in the instruction tuning of SEED-LLaMA.}
\label{tab:sft_data}
\begin{tblr}{
  cell{2}{1} = {r=4}{},
  cell{6}{1} = {r=2}{},
  cell{8}{1} = {r=3}{},
  cell{12}{1} = {r=3}{},
  cell{15}{1} = {r=7}{},
  cell{23}{1} = {r=5}{},
  vline{2} = {1-28}{},
  hline{1-2,5-6,7-8,11-12,15,22-23,28} = {-}{},
  hline{3-4,9-10,13-14,16-21,24-27} = {2-4}{},
}
Task                            & Dataset Name          & Dataset Description &   Type                                                                            \\
{Text-to-Image\\Generation   }   & {JourneyDB~\\\cite{pan2023journeydb}}        & {It contains 4429K Midjourney images, with text\\prompt, image caption, and QA pairs.} &Single-turn            \\
                                 & {DiffusionDB~\\\cite{wang2022diffusiondb}}     & {It contains 14 million images generated by Stable\\Diffusion using prompts by real users.} &Single-turn           \\
                                 & {LAION-Aesthetics\\~} & {It contains several collections of subsets from\\LAION 5B with high visual quality.} &Single-turn   \\
                                 & {VIST~\\\cite{huang2016visual}} & {It contains photos in 20K sequences, aligned to\\both caption and story language.} &Multi-turn               \\
{Multimodal\\Prompt Image\\ Generation}            & {Instructpix2pix\\\cite{brooks2023instructpix2pix}}  & {It contains text editing instructions and the\\corresponding images, with 454K samples.}   &Single-turn       \\
                                 & {MagicBrush~\\\cite{zhang2023magicbrush}}      & {It contains 10K manually annotated triplets\\(source image, instruction, target image).}    &Single-turn       \\
{Image\\Conversation} & {LLaVA~\\\cite{liu2023visual}}           & {We use 58K multi-turn conversations~between an \\assistant and a person.}       &Multi-turn                    \\
                                 & {SVIT~\\\cite{zhao2023svit}}             & {It contains conversations, complex reasoning,\\referring QA and detailed image description.}     &Multi-turn   \\
                                 & {LLaVAR~\\\cite{zhang2023llavar}}          & {It contains 16K multi-turn conversations, each\\with QA pairs for text-rich images.}     &Multi-turn           \\
{Multi-Image\\Understanding}  & {GSD~\\\cite{li2023mimic}}             & {It contains 141K pairs of images with text\\describing the differences.}   &Single-turn                         \\
{Image\\Captioning}                 & {VSR~\\\cite{liu2023visualspatial}}             & {It contains texts describing the spatial\\relations in the image, with 7K training samples.}  &Single-turn      \\
                                 & {COCO Caption~\\\cite{chen2015microsoft}}    & {It contains image-text pairs with human-written\\captions, with 82K training samples.}          &Single-turn    \\
                                 & {TextCaps~\\\cite{sidorov2020textcaps}}        & {It requires the model to comprehend and reason\\the text in images, with 21K training samples.}   &Single-turn  \\
Image QA                         & {VQAv2~\\\cite{goyal2017making}}            & {A dataset for open-ended image question \\answering, with 82K training samples.}                  &Single-turn  \\
                                 & {OKVQA~\\\cite{marino2019ok}}           & {It contains questions that require outside\\knowledge to answer, with 9K training samples.}       &Single-turn  \\
                                 & {A-OKVQA~\\\cite{schwenk2022okvqa}}         & {It is a successor of OKVQA containing more\\challenging questions, with 17K training samples.}    &Single-turn  \\
                                 & {GQA~\\\cite{hudson2019gqa}}             & {It contains questions for image understanding\\and reasoning, with 30K training samples.}    &Single-turn \\
                                 & {VizWiz~\\\cite{gurari2018vizwiz}}          & {It contains visual questions asked by people who\\are blind, with 20K~training samples.}          &Single-turn  \\
                                 & {TextVQA~\\\cite{singh2019towards}}         & {It contains questions that require models to read\\text in the image, with 800K training samples.} &Single-turn \\
                                 & {OCR-VQA~\\\cite{mishra2019ocr}}         & {It contains questions that requires reasoning about\\text to answer, with 173K training samples.}  &Single-turn \\
{Video\\Conversation}               & {Video-ChatGPT~\\\cite{maaz2023video}}   & {It contains of 100K video-instruction pairs\\via manual and semi-automated pipeline.}   &Single-turn   \\
Video QA                         & {ActivityNet~\\\cite{caba2015activitynet}}     & {It contains 200 different types of activities from\\YouTube, with 10K training videos.}            &Single-turn \\
                                 & {Next-QA~\\\cite{xiao2021next}}         & {It contains~52K QA pairs of videos grouped into\\causal, temporal and descriptive questions.}      &Single-turn \\
                                 & {MSVD~\\\cite{chen2011collecting}}             & {It contains videos~from YouTube with~descriptions,\\containing 1.2K training samples.}           &Single-turn   \\
                                 & {MSR-VTT~\\\cite{xu2016msr}}         & {It contains videos~from YouTube with~descriptions,\\containing 19K training samples.}             &Single-turn  \\
                                 & {iVQA~\\\cite{yang2021just}}            & {It is a video QA dataset with mitigated language\\biases, containing 6K training samples.} &Single-turn         
\end{tblr}
\end{table}

\begin{table}
\renewcommand\arraystretch{1.3}
\centering
\caption{Details of prompt templates used in supervised instruction tuning of SEED-LLaMA.}
\label{tab:sft_prompt}
\begin{tabular}{l|l} 
\hline
Type                                                                   & Prompt                                                                                                                                                                                 \\ 
\hline
Text-to-Image Generation                                               & USER: \{caption\} Please generation an image.\textbackslash{}nASSISTANT: \{image\}                                                                                                     \\ 
\hline
\begin{tabular}[c]{@{}l@{}}Multimodal Prompt\\Image Generation\end{tabular}                                                & \begin{tabular}[c]{@{}l@{}}USER: \{image1\} \{instruction\} Please generation an image.\\\textbackslash{}nASSISTANT: \{image2\}\end{tabular}                                           \\ 
\hline
\begin{tabular}[c]{@{}l@{}}Image Conversation\end{tabular} & USER: \{image\} \{question\}\textbackslash{}nASSISTANT: \{answer\}                                                                                                                     \\ 
\hline
\begin{tabular}[c]{@{}l@{}}Multi-Image\\Understanding\end{tabular}  & \begin{tabular}[c]{@{}l@{}}USER: This is the first image. \{image1\} This is the second image. \\\{image2\} \{question\}\textbackslash{}nASSISTANT: \{answer\}\end{tabular}            \\ 
\hline
Image Captioning                                                       & \begin{tabular}[c]{@{}l@{}}USER: \{image\} Please provide an accurate and concisedescription of\\the given image.\textbackslash{}nASSISTANT: \{caption\}\end{tabular}                  \\ 
\hline
Image QA                                                               & \begin{tabular}[c]{@{}l@{}}USER: \{image\} \{question\} Please provide an accurate answer consisting\\of only one word or phrase.\textbackslash{}nASSISTANT: \{answer\}\end{tabular}    \\ 
\hline
Video Conversation                                                     & USER: \{video\} \{question\}\textbackslash{}nASSISTANT: \{answer\}                                                                                                                     \\ 
\hline
Video QA                                                               & \begin{tabular}[c]{@{}l@{}}USER: \{video\} \{question\} Please provide an accurate answer~\\consisting of only one word or phrase.\textbackslash{}nASSISTANT: \{answer\}\end{tabular}  \\
\hline
\end{tabular}
\end{table}

\section{Evaluation}\label{sec:evaluation}
\subsection{Benchmarks}

\begin{table}[t]
\centering
\renewcommand{\arraystretch}{1.2}
\caption{Summary of the evaluation benchmarks.}
\resizebox{1.\linewidth}{!}{
\begin{tabular}{lllll}
\toprule
                        & Dataset                  & Task                      & Split         & Metric         \\
\midrule
\multirow{10}{*}{\rotatebox{90}{Image}} 
                        & COCO\cite{lin2014microsoft}      & Text-to-Image Generation & Karpathy test & CLIP score ($\uparrow$) \\
                        & Flickr30K~\cite{young2014image} & Text-to-Image Generation & test          & CLIP score ($\uparrow$) \\
                        & COCO Caption~\cite{chen2015microsoft}             & Scene Description         & test          & CIDEr ($\uparrow$)       \\
                        & VQAv2~\cite{goyal2017making}                    & Scene Understanding QA    & test-dev      & VQA acc. ($\uparrow$)   \\
                        & OKVQA~\cite{marino2019ok}                    & External Knowledge QA     & val           & VQA acc. ($\uparrow$)   \\
                        & VizWiz~\cite{gurari2018vizwiz}                   & Scene Understanding QA    & test-dev      & VQA acc. ($\uparrow$)   \\
                        & SEED-Bench~\cite{li2023seed}         & Comprehensive QA    & dim 1-9       & MCQ acc. ($\uparrow$) \\
\midrule
\multirow{4}{*}{\rotatebox{90}{Video}}  & MSVDQA~\cite{chen2011collecting}                   & Event Understanding QA    & test          & Top-1 acc. ($\uparrow$) \\
                        & MSRVTTQA~\cite{xu2016msr}                 & Event Understanding QA    & test          & Top-1 acc. ($\uparrow$) \\
                        & NExTQA~\cite{yang2021just}                   & Temporal/Causal QA        & test          & WUPS ($\uparrow$)       \\
\bottomrule
\end{tabular}}
\label{tab:benchs}
\end{table}

In order to assess the multimodal comprehension and image generation ability of SEED-LLaMA, we evaluate SEED-LLaMA on 10 benchmarks as shown in Tab.~\ref{tab:benchs}. For the evaluation of image generation, we adopt the CLIP-ViT-L/14 to calculate the CLIP score between the ground-truth image and the generated image. When evaluating SEED-Bench, we adhere to the official guidelines, selecting the option with the highest log likelihood as the response for each multi-choice question (MCQ). For the evaluation on video tasks, we uniformly sample 4 frames for MSVDQA and MSRVTTQA, and 8 frames for NExTQA. For the other tasks,  we follow the evaluation procedures in prior works~\cite{li2023blip,sun2023generative} and either submit the results to the official server (VQAv2, VizWiz) or assess them using the official code ourselves.

\subsection{Prompt Templates}
We summarize the prompt templates used for evaluating SEED-LLaMA in Tab. \ref{tab:comprehension_prompt}. As the pre-trained SEED-LLaMA with size of 8B and 14B adopt different LLM (Vicuna-7B and 
Llama2-chat-13B), their prompts differ accordingly.

\begin{table}
\centering
\renewcommand{\arraystretch}{1.2}
\setlength\tabcolsep{3pt}
\caption{Summary of the prompting template for evaluating SEED-LLaMA.}
\label{tab:comprehension_prompt}
\begin{tabular}{cll} 
\toprule
\multicolumn{1}{c}{Model}                                                                          & Type             & Template                                                                                                                                                                \\ \hline
\multirow{3}{*}{\begin{tabular}[c]{@{}c@{}}SEED-LLaMA\\8B\end{tabular}}                              & Image Captioning & \{image\}                                                                                                                                                               \\
                                                                                                   & Image QA         &\begin{tabular}[c]{@{}l@{}} \{image\}USER: \{question\} Please provide an accurate answer\\consisting of only one word or phrase.\textbackslash{}nASSISTANT:\end{tabular}                                        \\
                                                                                                   & Video QA         &\begin{tabular}[c]{@{}l@{}} \{video\}USER: \{question\} Please provide an accurate answer\\consisting of only one word or phrase.\textbackslash{}nASSISTANT:\end{tabular}                                        \\ \hline
\multirow{3}{*}{\begin{tabular}[c]{@{}c@{}}SEED-LLaMA\\14B\end{tabular}}                             & Image Caption    & \{image\}                                                                                                                                                               \\
                                                                                                   & Image QA         &\begin{tabular}[c]{@{}l@{}} \{image\} Please provide an accurate answer consisting of only\\one word or phrase based on the image.\textbackslash{}n \\Question:\{question\} \textbackslash{}n Answer:\end{tabular}  \\
                                                                                                   & Video QA         &\begin{tabular}[c]{@{}l@{}} \{video\} Please provide an accurate answer consisting of only\\one word or phrase based on the video.\textbackslash{}n \\Question:\{question\}\textbackslash{}n Answer:\end{tabular} \\ \hline
\multirow{3}{*}{\begin{tabular}[c]{@{}c@{}}SEED-LLaMA-I\\8B \& 14B\end{tabular}} & Image Caption    &\begin{tabular}[c]{@{}l@{}} USER: \{image\}Please provide an accurate and con-\\cise description of the given image.\textbackslash{}nASSISTANT:\end{tabular}                                           \\
\multicolumn{1}{l}{}                                                                               & Image QA         &\begin{tabular}[c]{@{}l@{}} USER: \{image\}\{question\} Please provide an accurate answer\\consisting of only one word or phrase.\textbackslash{}nASSISTANT:\end{tabular}                                         \\
\multicolumn{1}{l}{}                                                                               & Video QA         &\begin{tabular}[c]{@{}l@{}} USER: \{video\}\{question\} Please provide an accurate answer\\consisting of only one word or phrase.\textbackslash{}nASSISTANT:\end{tabular}                                         \\
\bottomrule
\end{tabular}
\end{table}

\section{Qualitative Cases}\label{sec:qualitative}
\begin{figure}
	\centering
	\includegraphics[width=1.0\linewidth]{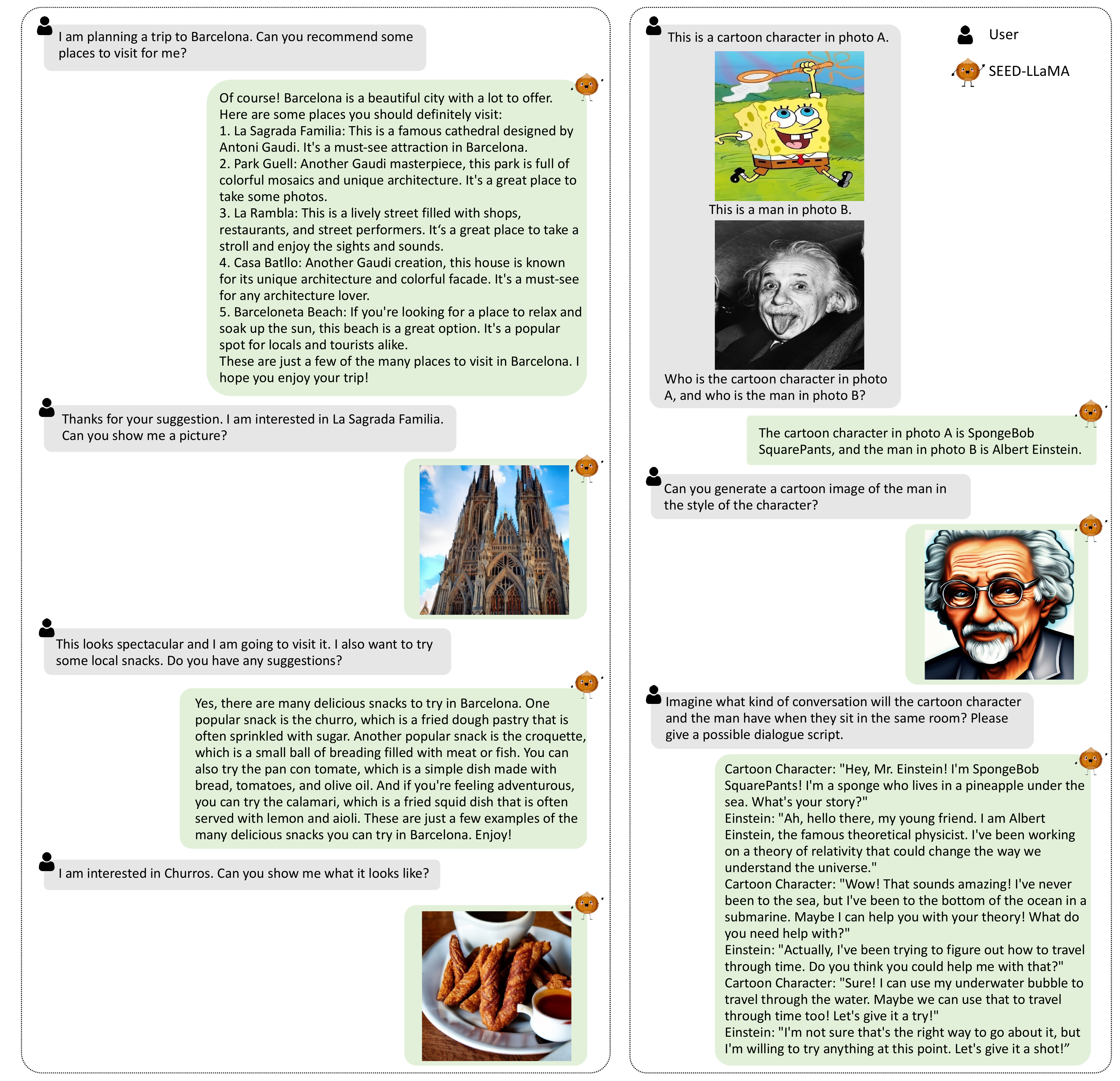}
\caption{Qualitative examples of multi-turn in-context image and text
generation by SEED-LLaMA given multimodal instructions.}
	\label{fig:multi_turn_appendix}
  \vspace{-10pt}
\end{figure}

\begin{figure}
	\centering
	\includegraphics[width=1.0\linewidth]{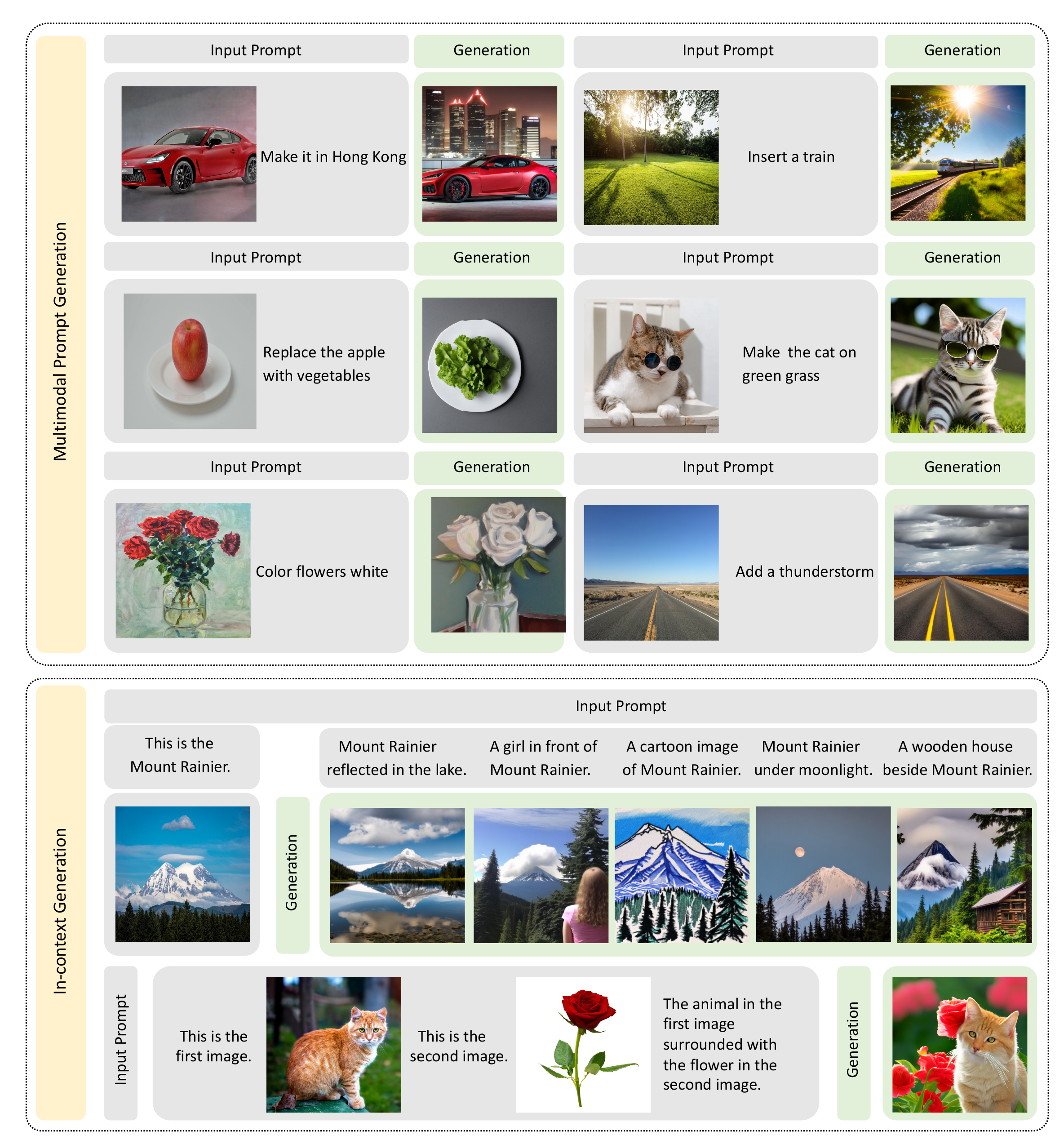}
\caption{Qualitative examples of compositional image generation by SEED-LLaMA.}
	\label{fig:results_appendix}
  \vspace{-10pt}
\end{figure}

\begin{figure}
	\centering
	\includegraphics[width=1.0\linewidth]{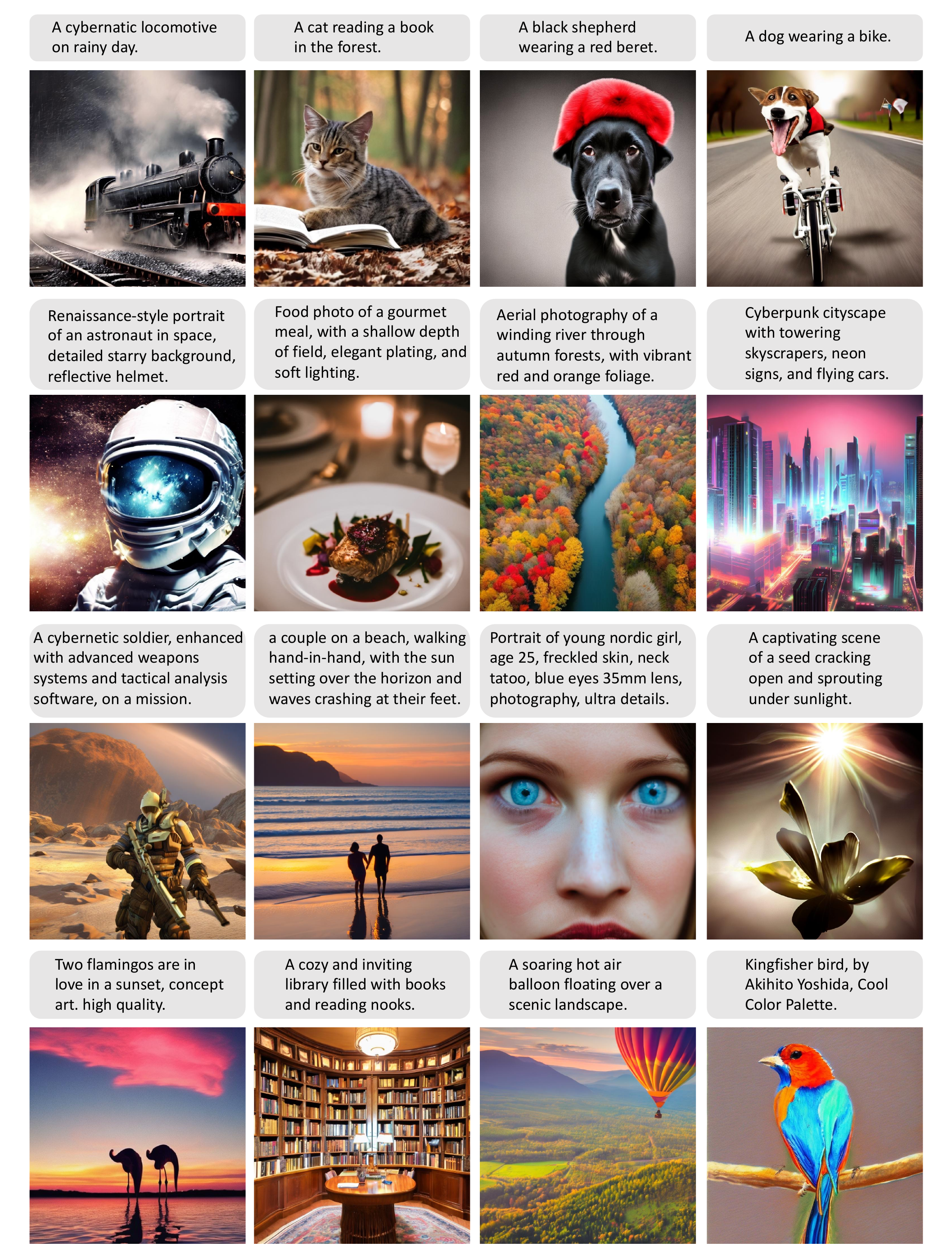}
\caption{Qualitative examples of text-to-image generation by SEED-LLaMA.}
	\label{fig:generation}
\end{figure}

\begin{figure}
	\centering
	\includegraphics[width=1.0\linewidth]{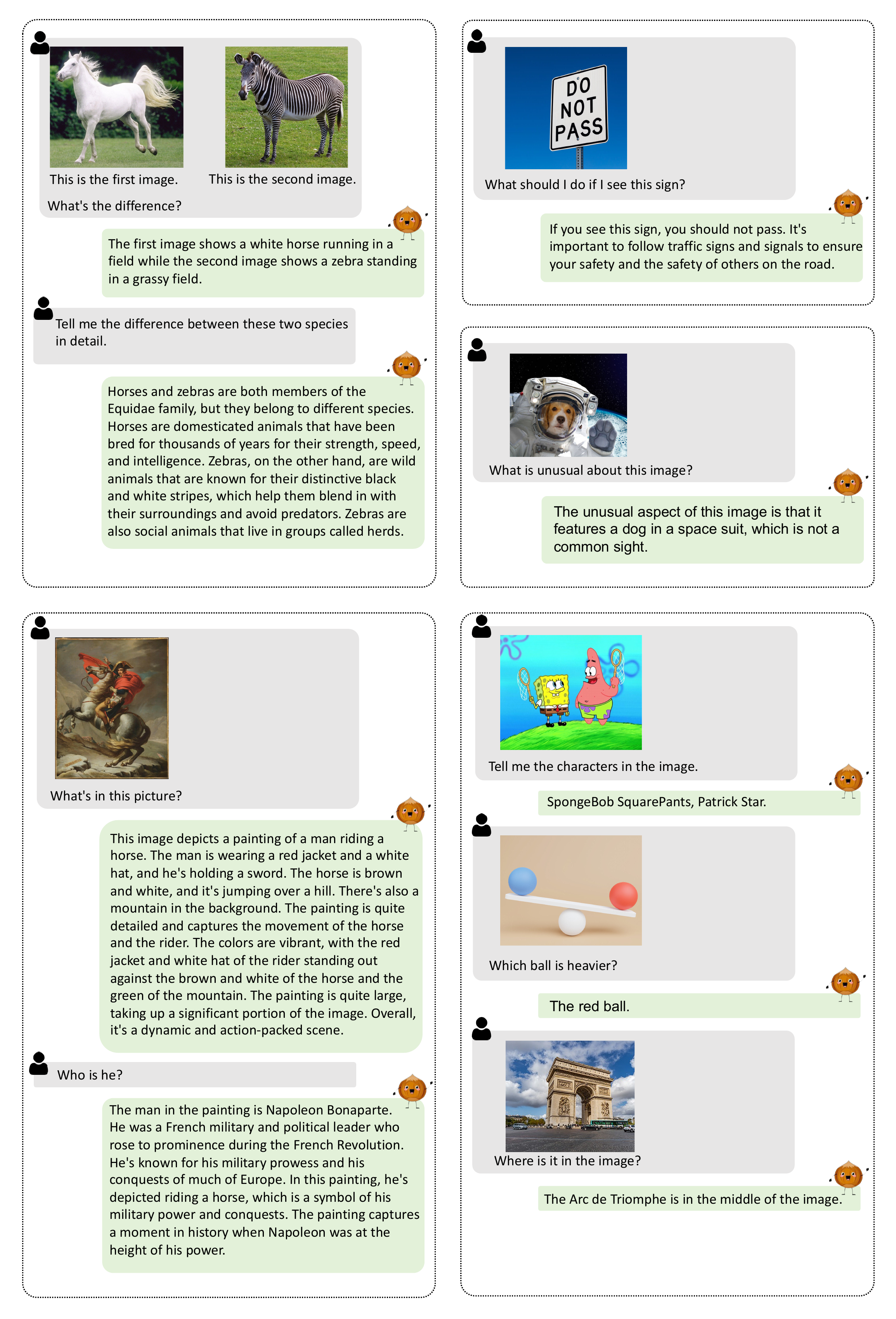}
\caption{Qualitative examples of multimodal comprehension by SEED-LLaMA.}
	\label{fig:image_comprehension_case_1}
\end{figure}

\begin{figure}
	\centering
	\includegraphics[width=1.0\linewidth]{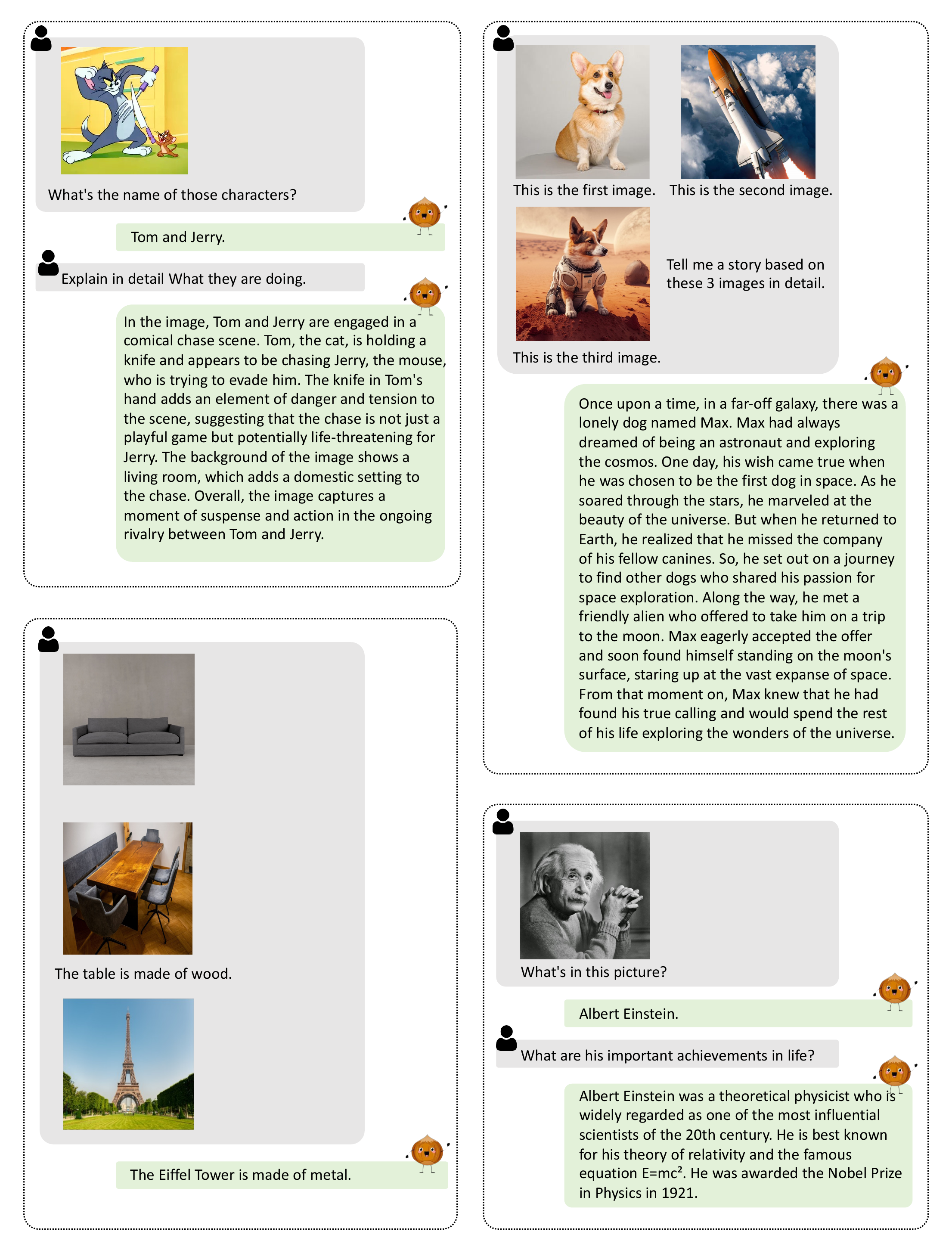}
\caption{Qualitative examples of multimodal comprehension by SEED-LLaMA.}
	\label{fig:image_comprehension_case_2}
\end{figure}

\begin{figure}
	\centering
	\includegraphics[width=1.0\linewidth]{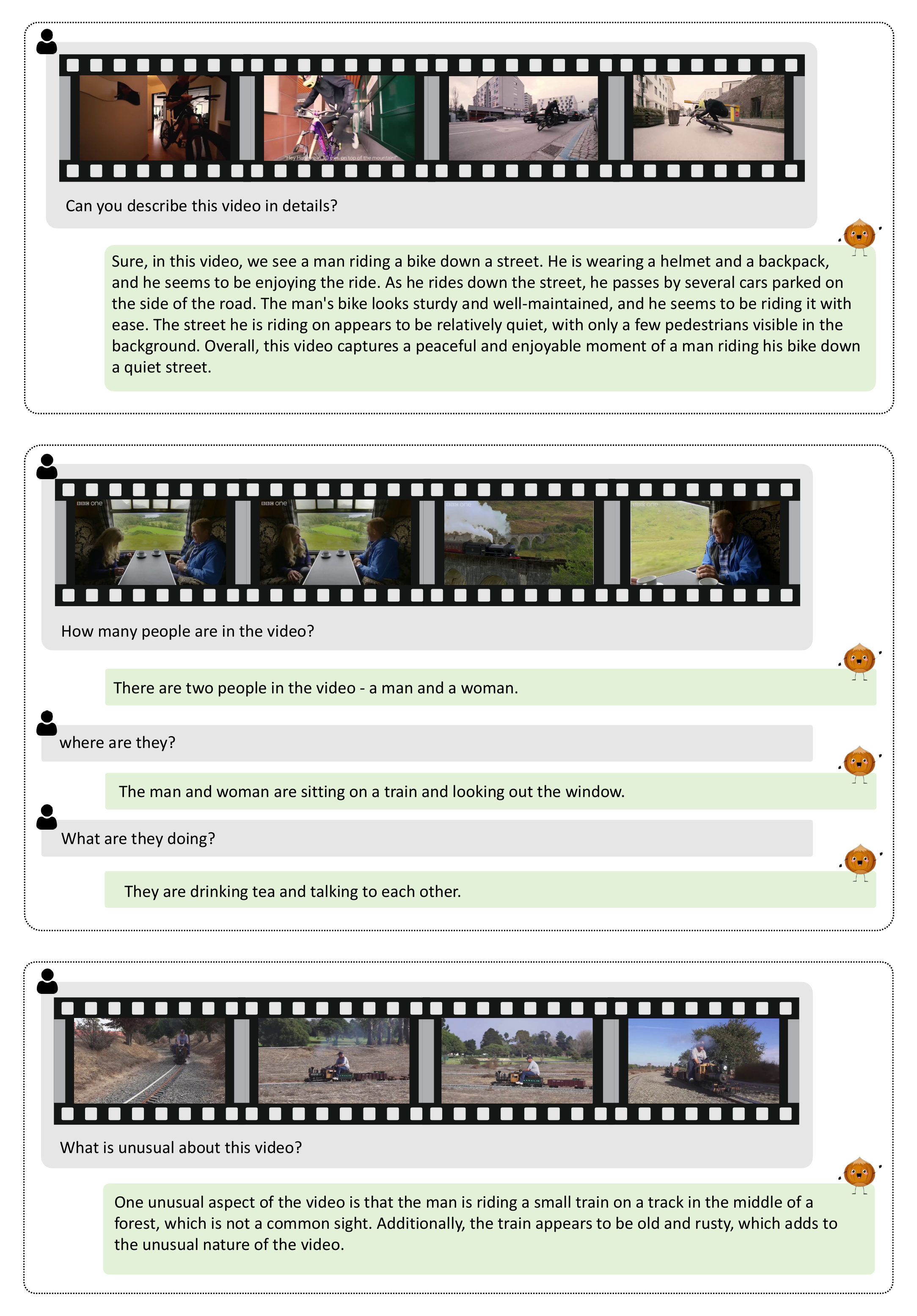}
\caption{Qualitative examples of multimodal comprehension by SEED-LLaMA.}
	\label{fig:video_comprehension_case}
\end{figure}

More examples of multi-turn in-context multimodal generation and compositional image generation are shown in Fig.~\ref{fig:multi_turn_appendix} and Fig.~\ref{fig:results_appendix}. Note that generating images with multimodal prompt is not an emergent ability since SEED-LLaMA is fine-tuned on corresponding paired data such as InstructPix2Pix~\cite{brooks2023instructpix2pix}.
We showcase qualitative examples of text-to-image generation by SEED-LLaMA in Fig.~\ref{fig:generation}. Given various textual descriptions, our SEED-LLaMA can generate realistic images that aligns with the text prompts. We further provide qualitative examples of multimodal comprehension by SEED-LLaMA in Fig.~\ref{fig:image_comprehension_case_1}, Fig.~\ref{fig:image_comprehension_case_2} and Fig.~\ref{fig:video_comprehension_case}. SEED-LLaMA can realize in-context multi-image understanding, real-world knowledge grounding, complex reasoning, story creation and video understanding.


\end{document}